\pdfoutput=1

\documentclass[11pt]{article}

\usepackage{tcolorbox}
\usepackage{booktabs}
\usepackage{amssymb} 
\usepackage{placeins} 
\usepackage[ruled,vlined,linesnumbered]{algorithm2e}
\SetKwInput{KwData}{Data}
\SetKwInput{KwResult}{Result}
\SetKwProg{Fn}{Function}{}{end}
\SetKwFunction{FChain}{Chain-of-Table}
\SetKwFunction{FDynamicPlan}{DynamicPlan}
\SetKwFunction{FGenerateArgs}{GenerateArgs}
\SetKwFunction{FQuery}{Query}
\usepackage{multirow}
\usepackage{makecell}

\usepackage[final]{acl}
\usepackage{amsmath}
\usepackage{enumitem}

\usepackage{times}
\usepackage{latexsym}
\usepackage{float} 
\usepackage[T1]{fontenc}
\usepackage{textcomp}

\usepackage{soul}

\usepackage[utf8]{inputenc}

\usepackage{microtype}

\graphicspath{./}

\usepackage{graphicx}

\title{\textsc{Chain-of-Query}: Unleashing the Power of LLMs in SQL-Aided Table Understanding via Multi-Agent Collaboration}

\author{
{\bfseries Songyuan Sui\textsuperscript{1}, 
Hongyi Liu\textsuperscript{1}, 
Serena Liu\textsuperscript{1}, 
Li Li\textsuperscript{2},} \\
{\bfseries Soo-Hyun Choi\textsuperscript{3}, 
Rui Chen\textsuperscript{2}, 
Xia Hu\textsuperscript{1}} \\
\textsuperscript{1}Rice University \\
\texttt{\{Songyuan.Sui, Hongyi.Liu, Serena.Liu, Xia.Hu\}@rice.edu} \\
\textsuperscript{2}Samsung Electronics America  \hfill \textsuperscript{3}Warner Bros. Discovery\\
\texttt{\{li.li1, rui.chen1\}@samsung.com} \hfill \texttt{soohyun.choi@wbd.com}
}

\begin{document}
\maketitle
\begin{abstract}
Table understanding requires structured, multi-step reasoning. Large Language Models (LLMs) struggle with it due to the structural complexity of tabular data. Recently, multi-agent frameworks for SQL generation have shown promise in tackling the challenges of understanding tabular data, but existing approaches often suffer from limitations such as the inability to comprehend table structure for reliable SQL generation, error propagation that results in invalid queries, and over-reliance on execution correctness. To address these issues, we propose \textsc{Chain-of-Query} (CoQ), a novel multi-agent framework for SQL-aided table understanding. CoQ adopts natural-language-style representations of table schemas to abstract away structural noise and enhance understanding. It employs a clause-by-clause SQL generation strategy to improve query quality and introduces a hybrid reasoning division that separates SQL-based mechanical reasoning from LLM-based logical inference, thereby reducing reliance on execution outcomes. Extensive experiments across four models and five widely used benchmarks demonstrate that CoQ achieves substantial accuracy improvements and significantly lowers invalid SQL rates compared to prior generic LLM-based, SQL-aided, and hybrid baselines, confirming its superior effectiveness in table understanding. The code is available at \url{https://github.com/SongyuanSui/ChainofQuery}.

\end{abstract}

\section{Introduction}

Large Language Models (LLMs) have shown remarkable performance across a variety of natural language processing (NLP) tasks \citep{yang2023harnessingpowerllmspractice}. However, they still struggle with understanding tabular data \citep{sui2023table}. This challenge stems from two main factors. First, the structure of tabular data differs significantly from that of plain text. It organizes information via rows and columns, introducing hierarchical relationships, positional dependencies, and implicit semantics that are challenging for language models to capture. Second, table-based tasks often involve multi-step, structured reasoning operations, such as aggregation, comparison, and arithmetic computation \citep{lu2025large}. They go beyond surface-level language understanding.

To adapt LLMs for table understanding, prior work has explored many approaches. Many studies \citep{Herzig_2020, zhang2023generative, zhang2023tablellama, li2024tablegpt, he2025tablelora} fine-tuned LLMs on table-specific datasets, but these methods treat the task as a single-turn generation problem, which results in shallow reasoning paths and overlooks intermediate data. Recent studies \citep{wang2024chaintab, ji2024treetab, zhou2025efficientmultiagentcollaborationtool} built agent-based pipelines with table manipulation tools to conduct multi-step reasoning. Nevertheless, these tools rely on human-defined programs whose functionalities are restricted and cannot perform complex or adaptive operations beyond their predefined scope.

\begin{figure*}[t]
  \centering
  \includegraphics[width=\linewidth]{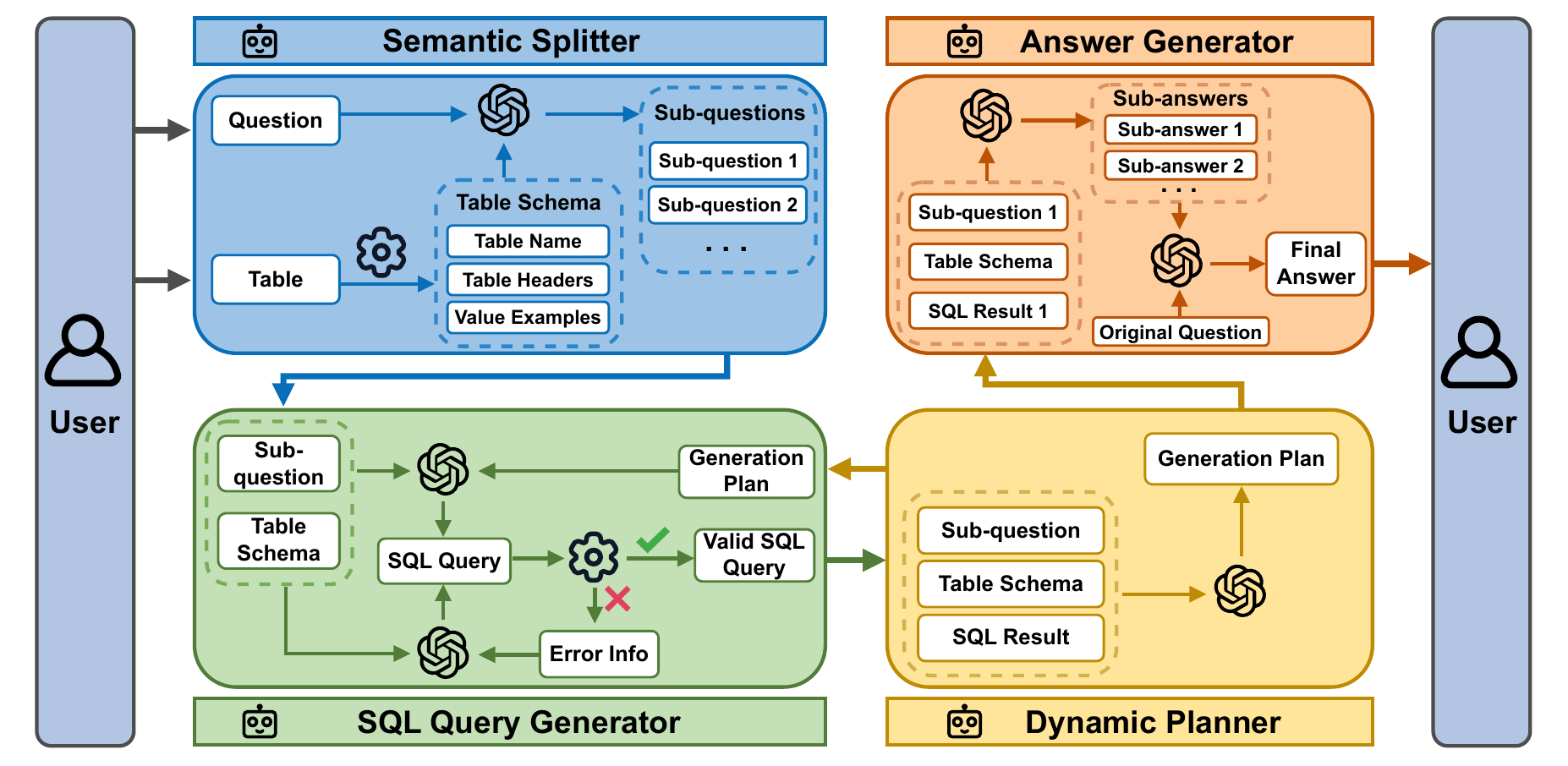}
  \caption{Overview of \textsc{Chain-of-Query} Framework.}
  \label{fig:overview}
\end{figure*}

We argue that incorporating Structured Query Language (SQL) provides a more principled approach for understanding tabular data. SQL is inherently designed to access, filter, and aggregate information from relational tables, making it well-aligned with the needs of table understanding. Although existing work on LLM-based SQL generation has made some progress in table understanding \citep{ye2023dater, kong2024opentab, huang2024queryagentreliableefficientreasoning, abhyankar2025hstarllmdrivenhybridsqltext}, its performance remains suboptimal. \textbf{We identify three key challenges in SQL-aided table understanding with LLMs.} \textbf{1.} Existing approaches typically feed the table and task instructions into an LLM, assuming the LLM can understand the table in order to produce meaningful SQL. However, if the LLM already struggles with table understanding, it is unlikely to generate accurate SQL in the first place. This workflow thus inherits LLMs' structural inability, as effective SQL generation still requires a solid grasp of the table’s layout and semantics—something LLMs inherently lack. \textbf{2.} Supporting complex, multi-step reasoning often requires SQL queries with multiple clauses and nested structures. Such queries are difficult to construct correctly in a single pass, and even small errors can cascade into query failure. \textbf{3.} Existing methods depend heavily on the correctness of SQL execution, as they directly use SQL results as final answers. This further leads LLMs to generate overly complicated queries, thereby reducing overall reliability.

To tackle these challenges, we propose \textsc{Chain-of-Query} (CoQ), a novel multi-agent framework for SQL-aided table understanding. To address the first challenge, CoQ replaces raw table inputs with \textbf{natural-language representations of table schemas}. Our insight is that SQL generation depends primarily on high-level schemas, which can be easily and accurately expressed in natural language, bypassing the need for LLMs to interpret complex tabular structures. This abstraction allows LLMs to focus on semantics without being hindered by noisy or irregular layouts. To address the second challenge, we propose a \textbf{Clause-by-Clause SQL Generation Strategy}, which incrementally constructs SQL queries one clause at a time, forming a chain of progressively refined queries. To solve the third challenge, we design a \textbf{Hybrid Reasoning Division Strategy} that separates mechanical reasoning (executed via SQL) from logical reasoning (delegated to LLMs), treating SQL outputs as intermediate steps rather than final answers. In addition, our \textbf{Parallel Decomposition} enhances robustness by splitting complex questions into parallelizable sub-questions, avoiding the inter-step dependencies of traditional sequential reasoning. Together, CoQ offers a novel design that directly addresses these challenges, establishing an effective new paradigm for table understanding.

To comprehensively evaluate CoQ, we conduct extensive experiments with both closed-source models (GPT-3.5, GPT-4.1) and open-source models (LLaMA 2, DeepSeek-V3) on five representative benchmarks: WikiTQ, TabFact, FeTaQA, IM-TQA, and Open-WikiTable. CoQ is compared with strong baselines across four categories: generic LLM-based, SQL-aided, hybrid, and general structured data reasoning. Results show that CoQ consistently outperforms state-of-the-art (SOTA) methods; for example, on WikiTQ it improves accuracy from 61.11\% to 74.77\% and reduces invalid-SQL rates from 9.48\% to 3.34\% with GPT-3.5.

In summary, we propose \textsc{Chain-of-Query} (CoQ), a novel multi-agent collaboration framework for SQL-aided table understanding with the following contributions:

\begin{itemize}[noitemsep, topsep=1pt]
    \item We design natural-language-style table schemas to abstract away structural noise and irregularities in raw tables, enabling LLMs to focus on high-level semantics.
    \item We propose a novel Clause-by-Clause SQL Generation Strategy, which constructs queries incrementally to reduce error propagation and improve reliability.
    \item We introduce a Hybrid Reasoning Division Strategy that separates mechanical reasoning (handled by SQL) and logical reasoning (handled by LLMs). 
    \item Extensive experiments show that \textsc{Chain-of-Query} consistently outperforms previous baselines, achieving substantial improvements across diverse settings.
\end{itemize}

\section{Related Work}

We review related work in three areas: LLM-based table understanding, Text-to-SQL, and hybrid methods for table reasoning.
Additional discussions on multi-agent approaches for table understanding and general structured data reasoning with LLMs can be found in Appendix~\ref{sec:appendix_more_related}.

\noindent\textbf{LLM-Based Table Understanding.} Since the structure of tables (e.g., non-sequential cell order) differs significantly from plain text typically used in pre-training \citep{raffel2023exploringlimitstransferlearning}, several studies \citep{zhang2023generative, zhang2023tablellama, li2024tablegpt, zhuang2024structlmbuildinggeneralistmodels, he2025tablelora} fine-tuned LLMs to adapt to tabular data. While effective on specific datasets, these methods depend on manually crafted instructions and expensive training, limiting scalability. More recent prompt-based approaches \citep{sui2023table, sui2023tap4llm, chen2024tablerag} exploit LLMs’ general reasoning ability, offering better generalization across tasks. However, these methods typically adopt single-turn prompting strategies, failing to support multi-hop reasoning.


\noindent\textbf{Text-to-SQL.} Generating SQL queries from natural language is a long-standing challenge. Early methods primarily relied on fine-tuning LLMs on annotated SQL datasets \citep{pourreza2024dtssqldecomposedtexttosqlsmall, li2024codesbuildingopensourcelanguage}. To reduce training cost and improve compositional reasoning, \citet{tai2023exploringchainofthoughtstyleprompting} and \citet{pourreza2023dinsqldecomposedincontextlearning} explored Chain-of-Thought prompting in Text-to-SQL. OpenTab \citep{kong2024opentab} integrates LLM-based SQL generation into a tabular data RAG system. Recently, MAC-SQL \citep{wang2025macsql} and MAG-SQL \citep{xie2024mag} build multi-agent frameworks that refine generated queries to improve SQL quality. Although these methods perform well on benchmarks such as Spider \citep{yu2019spiderlargescalehumanlabeleddataset} and BIRD \citep{li2023llmservedatabaseinterface}, they all tend to generate fully grounded SQL queries that directly return the final answer, making query construction difficult. Moreover, LLMs tend to produce overly complex queries in an attempt to handle questions holistically \citep{shen2025studyincontextlearningbasedtexttosqlerrors}, reducing reliability.

\noindent\textbf{Hybrid Methods for Table Understanding.} To address the shortcomings of LLMs in structured reasoning, hybrid approaches integrate SQL with LLMs. ReAcTable \citep{zhang2023reactableenhancingreacttable} generates multiple SQL candidates for intermediate information retrieval but at the cost of increased computation and limited structural understanding. SynTQA \citep{zhang2024syntqasynergistictablebasedquestion} generates both Text-to-SQL and LLM answers but only selects one result, discarding complementary information. Recent studies \citep{nahid2024tabsqlifyenhancingreasoningcapabilities, cao2025tablemasterrecipeadvancetable} use SQL to select key columns and rows before invoking LLMs. While effective for data reduction, they apply SQL only at a basic level, overlooking its potential for complex table manipulations. H-STAR \citep{abhyankar2025hstarllmdrivenhybridsqltext} further employs complex SQL-based symbolic reasoning, but its reliance on single-shot SQL generation at each stage makes it vulnerable to execution failures.

\section{\textsc{Chain-of-Query} Approach}

\subsection{Problem Definition of Table Understanding}

Given a 2-tuple $X = (Q, T)$, where $Q$ is a natural language question and $T = \{S, D\}$ is a table composed of schema $S$ and data content $D$, the goal of table understanding is to identify a relevant subset of data $D_r \subseteq D$ that is necessary to answer $Q$, and then derive the final answer $A$ by reasoning over both $Q$ and $D_r$. Here, the schema $S$ describes the structure of the table, while $D$ contains the individual cell-level values.

\subsection{Overview of \textsc{Chain-of-Query}}

We propose \textsc{Chain-of-Query} as a multi-agent framework that decomposes table understanding into modular sub-tasks, with each agent invoking dedicated strategies for its assigned role. As shown in Figure~\ref{fig:overview}, it comprises four specialized agents: \textbf{Semantic Splitter}, \textbf{SQL Query Generator}, \textbf{Dynamic Planner}, and \textbf{Answer Generator}, each responsible for a distinct stage of the pipeline. The Semantic Splitter constructs natural-language-style table schemas and decomposes questions into parallel sub-questions. The SQL Query Generator applies the Clause-by-Clause SQL Generation Strategy. The Dynamic Planner incorporates the Hybrid Reasoning Division Strategy. The Answer Generator synthesizes final answers based on SQL outputs and inferences from the LLM. We describe each component and its corresponding strategies in the following sections.

\subsection{Semantic Splitter: Schema Abstraction and Query Decomposition}

The Semantic Splitter serves as the entry point of the entire agentic workflow. This module is designed to optimize the input for SQL generation by addressing two key challenges: (1) LLMs’ limited ability to interpret complex table structures, which often leads to inaccurate SQL generation; and (2) interference between sub-questions within a single complex query, where independent sub-questions could be handled separately to avoid entanglement. To address these challenges, we introduce a natural-language-style representation of table schemas for query generation, and incorporate a parallel decomposition mechanism to identify and isolate separable sub-queries for clean execution. The following subsections detail this process. An illustrative example is provided in Appendix~\ref{sec:appendix_decompose_example}.

\begin{figure}[htbp]
  \centering
  \includegraphics[width=\linewidth]{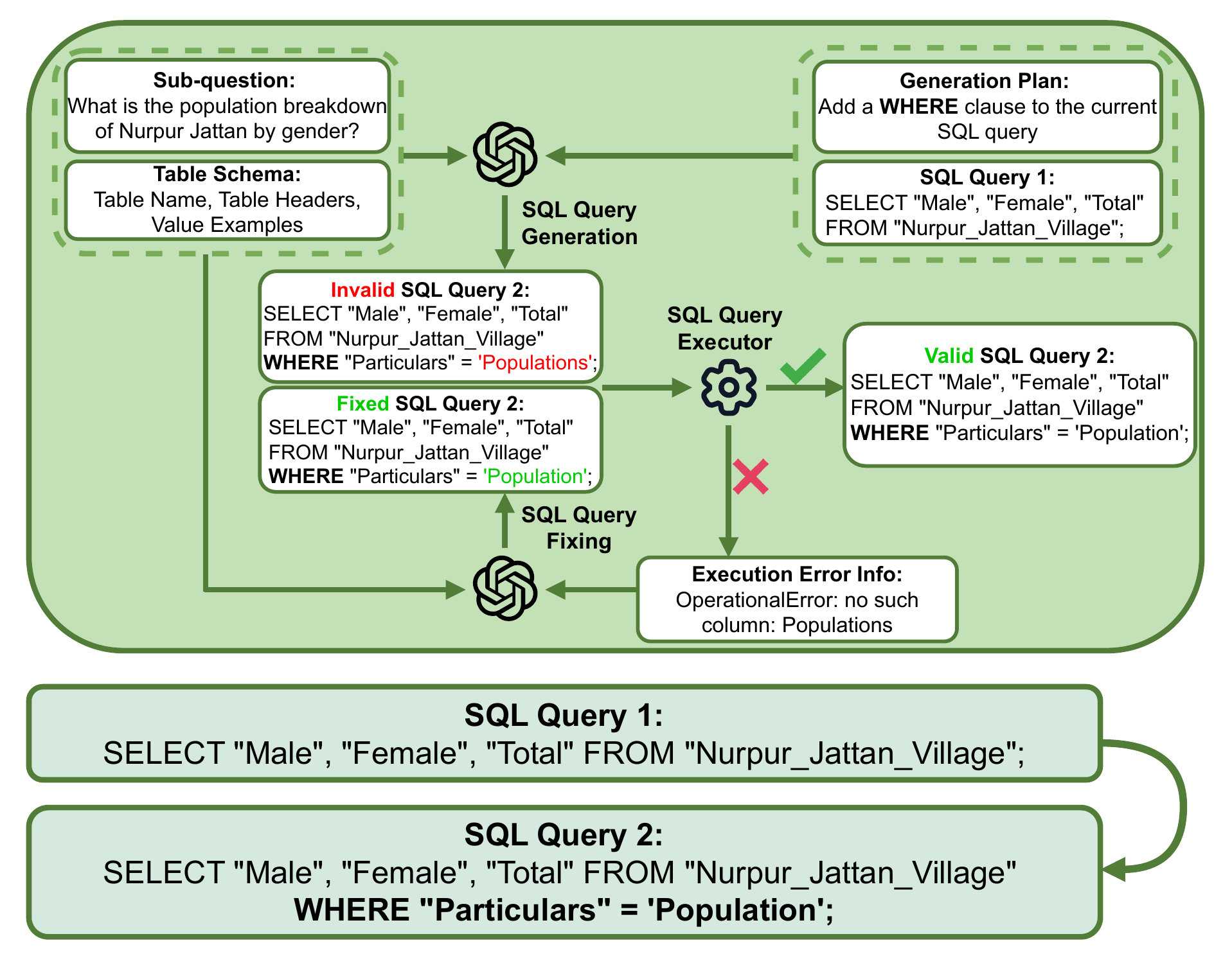}
  \caption{Illustration of the SQL Query Generator and the Clause-by-Clause Generation Strategy.}
  \label{fig:SQL_generator}
\end{figure}

\subsubsection{Natural-Language-Style Table Schema}

LLMs’ limited understanding of tabular data hinders their ability to generate accurate SQL based on raw tables. To address this, we ask: Is a natural-language-style schema sufficient for SQL generation? Our key insight is that SQL generation primarily depends on high-level schema information—not the detailed table contents—and that expressing the schema in natural language aligns better with LLMs’ strengths as language models.

To address these limitations in handling raw and irregular table structures, we construct a concise natural-language-style schema as a surrogate for full-table input. It consists of three components: \emph{table name}, \emph{headers}, and \emph{value examples}. The table name employs concise keywords to summarize the main content and purpose of the table, providing essential context for the LLM to understand the table. The table headers comprise multiple (column name, column value type) pairs. These pairs not only provide a natural-language abstraction of the table structure, but also serve as a reference for value formatting in downstream SQL generation. The value examples are a few sampled rows of the table paired with their corresponding column names. We adopt "PIPE" format (i.e., a pipe-delimited string with "|" separators) to encode the value examples, as studies found it to improve LLM performance on table tasks \citep{sui2023table, wang2024chaintab}.

By replacing the full table with our designed schema, LLMs can generate accurate SQL without being distracted by complex layouts or unnecessary content. This approach also improves scalability by avoiding full-table imports, thereby reducing input length and simplifying reasoning.

\subsubsection{Parallel Decomposition}

The second component of the Semantic Splitter is a Parallel Decomposer, which focuses on separating input questions into independent sub-questions. Unlike traditional sequential decomposition strategies, where each sub-question depends on the previous answer and becomes fragile due to error propagation, our approach focuses on questions without interdependencies. We defer the handling of sequential dependencies to our Clause-by-Clause Generator in Section~\ref{subsec:clause-gen}.

Many complex tabular questions involve comparing or aggregating information from semantically disjoint table regions, which do not require sequential reasoning. To address this, the Parallel Decomposer splits such questions into independent sub-questions, each targeting a localized region of the table. This removes inter-step dependencies and enables concurrent processing. Each sub-query then operates over a narrow, self-contained scope without interference from others, simplifying SQL generation.

\subsection{Clause-wise SQL Generation for Precision and Robustness}

After outlining our input transformation and preprocessing steps, we now focus on the next key component—generating accurate SQL with LLMs to support effective table understanding. The SQL Query Generator constructs executable SQL queries to retrieve intermediate information relevant to the task and support mechanical reasoning. As shown in Figure~\ref{fig:SQL_generator}, an initial query is generated by an LLM, then validated before being passed to the Dynamic Planner, which determines whether further clauses should be appended to improve precision. The following subsection details the generation process.

\subsubsection{SQL Query Generation with Validation}

Given the schema and sub-question, an SQL query is first generated by an LLM to retrieve relevant information. The query is validated by directly executing it via an SQL executor; if it fails, the error message is passed to a secondary LLM for correction. The revised query is re-executed and, if successful, forwarded to the Dynamic Planner. To enhance query quality, we introduce a clause-by-clause generation strategy described below.


\subsubsection{Clause-by-Clause SQL Generation}
\label{subsec:clause-gen}

Table understanding tasks often require complex, multi-step reasoning, such as aggregation and mathematical operations that in turn demand SQL queries with multiple clauses or even nested structures. This makes single-shot SQL generation particularly challenging, as even minor errors can propagate and cascade through the query, ultimately leading to execution failures.

To improve the reliability and controllability of SQL generation, we propose a Clause-by-Clause Strategy that incrementally builds queries by isolating the construction of each clause. The generator first produces a basic \texttt{SELECT-FROM} query selecting columns relevant to the sub-question. Without condition filtering, however, this query may return a large number of rows, posing scalability challenges. To mitigate this, we sample a few representative rows (similar to the natural-language-style table schema construction) as the Planner only needs to observe data patterns rather than precise values. 

Subsequent clauses (e.g., \texttt{WHERE}, \texttt{GROUP BY}) are appended one at a time, under the guidance of the Dynamic Planner. By decomposing SQL construction into discrete clauses, each step focuses solely on its own logic and syntax, independent of the full query context. Each intermediate query is validated before proceeding. If execution fails, the error can be traced to the most recently added clause, allowing for more targeted correction. If correction fails, the agent reverts to the last validated query, ensuring robustness. As shown in Figure~\ref{fig:SQL_generator}, if a newly generated clause introduces an error to the query (SQL2) and cannot be resolved, the agent uses the last valid query (SQL1) to ensure executable SQL. This incremental process forms a chain of increasingly precise queries, where each step builds on a validated state, preventing error propagation and ensuring stability. The list of clauses is provided in Appendix~\ref{sec:appendix_clause}.

\subsection{Balancing SQL and LLM Reasoning through Dynamic Planning}

The Dynamic Planner incrementally checks each newly generated query and its execution result to decide whether the current sub-table is sufficient for LLM reasoning. If the sub-table still contains irrelevant or indirect information, the Planner selects another SQL clause and asks the SQL Generator to append it to the current query. The Planner is guided by a Sufficiency-based Early Stopping Mechanism. Together with the Answer Generator, it jointly implements the Hybrid Reasoning Division Strategy: SQL handles mechanical reasoning, while LLMs perform logical reasoning over intermediate results to produce the final answer. 

\subsubsection{Hybrid Reasoning Division Strategy}

Existing SQL-aided approaches treat the SQL output as the final answer, requiring highly precise queries. This often leads to long, fragile SQL programs, especially when reasoning is complex or question intent is ambiguous. However, table understanding is inherently open-ended: the goal is to derive a natural language answer, not merely to retrieve an exact value. Based on these insights, we introduce a Hybrid Reasoning Division Strategy that decomposes table understanding into two stages: mechanical reasoning (e.g., filtering, arithmetic) is offloaded to SQL, whereas logical reasoning (e.g., comparison, inference) is handled by LLMs. Crucially, SQL execution results are treated as intermediate data rather than final answers, enabling LLMs to apply their stronger inference capabilities over SQL outputs to arrive at the final answer. This division leverages the respective strengths of SQL and LLMs. Our case study (Appendix~\ref{sec:appendix_case}) and examples (Appendix~\ref{sec:appendix_planner_example}) further explain this design.

\begin{table*}[ht]
\caption{Main results on WikiTQ and TabFact. \textbf{Acc.} = accuracy (\%), \textbf{Inv.} = invalid SQL rate (\%, lower is better). \underline{Underline} = second-best, \textbf{bold} = best. Improvements are over second-best. CoQ achieves the highest accuracy with a large margin, driven by schema abstraction, clause-by-clause SQL generation, and hybrid reasoning.}
\centering
\renewcommand{\arraystretch}{0.90}
\setlength{\tabcolsep}{2.0 pt}
\begin{tabular}{l*{8}{c}}
\toprule
\multirow{3}{*}{\textbf{Method}\rule{0pt}{9pt}} &
\multicolumn{4}{c}{\textbf{WikiTQ}\rule{0pt}{9pt}} &
\multicolumn{4}{c}{\textbf{TabFact}\rule{0pt}{9pt}} \\
\cmidrule(lr){2-5} \cmidrule(lr){6-9}
& \multicolumn{2}{c}{\textbf{GPT-3.5}\rule{0pt}{8pt}} & \multicolumn{2}{c}{\textbf{LLaMA 2}\rule{0pt}{8pt}} 
& \multicolumn{2}{c}{\textbf{GPT-3.5}\rule{0pt}{8pt}} & \multicolumn{2}{c}{\textbf{LLaMA 2}\rule{0pt}{8pt}} \\
& Acc. & Inv. & Acc. & Inv. & Acc. & Inv. & Acc. & Inv. \\
\midrule
\emph{Generic LLM-based Table Understanding} \\
\quad End-to-End QA & 43.39 & N/A & 35.48 & N/A & 67.45 & N/A & 53.46 & N/A  \\
\quad Table-to-Text & 16.07 & N/A & 14.21 & N/A & 49.72 & N/A & 48.06 & N/A  \\
\quad Few-Shot QA  & 52.56 & N/A & 35.52 & N/A & 71.54 & N/A & 62.01 & N/A \\
\quad Chain-of-Thought & 53.48 & N/A & 36.05 & N/A & 65.37 & N/A & 60.52 & N/A \\
\quad Binder & 56.74 & N/A & 30.92 & N/A & 79.17 & N/A & 62.76 & N/A \\
\quad Dater  & 52.81 & N/A & 41.44 & N/A & 78.01 & N/A & 65.12 & N/A \\
\quad Chain-of-Table  & 59.94 & N/A & 42.61 & N/A & 80.20 & N/A & 67.24 & N/A \\
\quad Tree-of-Table & \underline{61.11} & N/A & \underline{44.01} & N/A & \underline{81.92} & N/A & \underline{69.33} & N/A\\
\hline
\emph{SQL-aided Table Understanding} \\
\quad Basic Text-to-SQL & 47.40 & 14.07 & 32.18 & 18.81 & 64.93 & 19.66 & 63.27 & 33.02 \\
\quad OpenTab & 55.39 & 10.24 & 37.41 & \underline{15.16} & 78.57 & \underline{12.73} & 56.84 & \underline{27.83}  \\
\quad MAC-SQL& 52.92 & 10.34 & 36.88 & 17.65 & 76.06 & 18.13 & 56.68 & 31.54 \\
\quad MAG-SQL & 55.87 & \underline{9.48} & 38.25 & 16.13 & 78.84 & 13.28 & 59.91 & 29.49  \\[1pt] 
\hline \\[-10pt]
\renewcommand{\cellset}{\renewcommand{\arraystretch}{0.7}}
\quad \raisebox{0.6ex}[0pt]{\textsc{Chain-of-Query} (Ours)} &
\shortstack[c]{\textbf{74.77} \\[-1pt] \footnotesize\textcolor{green!40!black}{\textbf{(+13.66)}}} &
\shortstack[c]{\textbf{3.34} \\[-1pt] \footnotesize\textcolor{green!40!black}{\textbf{(-6.14)}}} &
\shortstack[c]{\textbf{58.91} \\[-1pt] \footnotesize\textcolor{green!40!black}{\textbf{(+14.90)}}} &
\shortstack[c]{\textbf{13.18} \\[-1pt] \footnotesize\textcolor{green!40!black}{\textbf{(-1.98)}}} &
\shortstack[c]{\textbf{92.31} \\[-1pt] \footnotesize\textcolor{green!40!black}{\textbf{(+10.39)}}} &
\shortstack[c]{\textbf{2.74} \\[-1pt] \footnotesize\textcolor{green!40!black}{\textbf{(-9.99)}}} &
\shortstack[c]{\textbf{78.80} \\[-1pt] \footnotesize\textcolor{green!40!black}{\textbf{(+9.47)}}} &
\shortstack[c]{\textbf{23.88} \\[-1pt] \footnotesize\textcolor{green!40!black}{\textbf{(-3.95)}}} \\
\Xhline{1.2pt}
\end{tabular}
\label{tab:performance_wiki_fact}
\end{table*}

\subsubsection{Sufficiency-based Early Stopping}

To support this hybrid reasoning strategy, we introduce the Sufficiency-based Early Stopping Mechanism. It enables the Planner to halt clause generation once the retrieved data are sufficient. This prevents unnecessary SQL complexity and ensures that reasoning responsibilities are dynamically and appropriately balanced between SQL and LLMs. The detailed algorithm is provided in Appendix~\ref{sec:appendix_stopping}.

\subsection{Answer Generator}

The Answer Generator produces the final answer by aggregating sub-answers from each sub-question. It first generates a sub-answer based on the corresponding SQL result, then combines all sub-answers into a complete natural language response to the original question. As the final stage of the pipeline, this component ensures that localized reasoning results are coherently integrated into a unified natural language response.

\section{Experiments}
In this section, we empirically evaluate the effectiveness of CoQ and aim to answer the following questions:  
\textbf{RQ1.} How does CoQ compare with popular table understanding methods (e.g., generic, SQL-based, and hybrid)?  
\textbf{RQ2.} How well does CoQ generalize to real-world, structurally complex tabular workloads?
\textbf{RQ3.} How does CoQ perform relative to existing methods in terms of cost?

\subsection{Experimental Setup}

\noindent\textbf{Datasets and Evaluation Metrics.} We evaluate our CoQ framework on five widely used table understanding benchmarks: WikiTQ \citep{pasupat2015wikitab}, FeTaQA \citep{Nan2021FeTaQAFT}, TabFact \citep{2019TabFactA}, IM-TQA \citep{zheng-etal-2023-im}, and Open-WikiTable \citep{kweon2023openwikitabledatasetopendomain}. WikiTQ and FeTaQA are table QA datasets requiring short-span and free-form answers, respectively. TabFact is a fact verification task based on tables. IM-TQA features complex real-world table styles, while Open-WikiTable involves multi-table scenarios. Collectively, these datasets cover diverse reasoning types, table structures, and domains, forming a robust testbed for evaluating generalization. Dataset details are provided in Appendix~\ref{sec:appendix_data}.

We use official accuracy metrics for WikiTQ, IM-TQA, and Open-WikiTable, and standard binary accuracy for TabFact. For FeTaQA, we report BLEU \citep{papineni2002bleu} and ROUGE \citep{lin2004rouge} scores to align with prior work. We also track the invalid SQL generation rate.

\noindent\textbf{Baselines.} We first compare our method against two categories of baseline approaches: (a) \emph{Generic LLM-based table understanding}, including End-to-End QA, Few-Shot QA, Table-to-Text \citep{min2024exploringimpacttabletotextmethods}, Chain-of-Thought \citep{wei2022cot}, Binder \citep{Binder}, Dater \citep{ye2023dater}, Chain-of-Table \citep{wang2024chaintab}, and Tree-of-Table \citep{ji2024treetab}; (b) \emph{SQL-aided table understanding}, including Basic Text-to-SQL \citep{rajkumar2022evaluating}, OpenTab \citep{kong2024opentab}, MAC-SQL \citep{wang2025macsql}, and MAG-SQL \citep{xie2024mag}. We further compare our CoQ with two more categories of baselines: (c) \emph{Hybrid table understanding}, including TabSQLify \citep{nahid2024tabsqlifyenhancingreasoningcapabilities}, SynTQA \citep{zhang2024syntqasynergistictablebasedquestion}, TableMaster \citep{cao2025tablemasterrecipeadvancetable}, and H-STAR \citep{abhyankar2025hstarllmdrivenhybridsqltext}; (d) \emph{General structured data reasoning}, including StructGPT \citep{jiang2023structgptgeneralframeworklarge}, QueryAgent \citep{huang2024queryagentreliableefficientreasoning}, Readi \citep{cheng2024necessaryllmsefficientlyfaithfully}, and Chain-of-Knowledge \citep{li2024chainofknowledgegroundinglargelanguage}. Results and analysis for category (d) are provided in Appendix~\ref{sec:appendix_more_experiments_structure} due to page limitations.

\noindent\textbf{Implementation Details.} To align with baselines, we adopt GPT-3.5 and LLaMA~2 as the backbone LLMs, and additionally evaluate with the recent stronger models LLaMA~3.1, DeepSeek-V3, and GPT-4.1. Model configurations are detailed in Appendix~\ref{sec:appendix_parameters}. Prompts include few-shot examples sampled from the training set, with illustrative cases shown in Appendix~\ref{sec:appendix_prompts_coq}.

\begin{table}[ht]
\caption{FeTaQA results with GPT-3.5. \textbf{BLEU} and \textbf{ROUGE} (R-1/2/L) evaluate answer quality. CoQ achieves the best results across BLEU, ROUGE, and error rate, enabled by its schema abstraction, clause-by-clause SQL generation, and hybrid reasoning.}
\centering
\renewcommand{\arraystretch}{0.92}
\setlength{\tabcolsep}{2.8pt} 
\begin{tabular}{lccccc}
\toprule
\textbf{Method} & \textbf{BLEU} & \textbf{R-1} & \textbf{R-2} & \textbf{R-L} & \textbf{Inv.} \\
\midrule
E2E QA & 16.94 & 0.60 & 0.38 & 0.50 & N/A \\
Tab2Text & 9.43 & 0.40 & 0.22 & 0.33 & N/A \\
CoTab & \underline{20.45} & \underline{0.62} & \underline{0.40} & \underline{0.52} & N/A \\
BT2SQL & 16.92 & 0.60 & 0.37 & 0.49 & 14.23 \\
OpenTab & 18.19 & 0.60 & 0.37 & 0.49 & \underline{10.71} \\
MAC-SQL & 17.56 & 0.58 & 0.35 & 0.47 & 13.98 \\
MAG-SQL & 18.81 & 0.60 & 0.37 & 0.48 & 11.13 \\
\midrule
\\[-8pt]
\raisebox{0.9ex}[0pt]{\textbf{CoQ} (Ours)} &
\shortstack[c]{\textbf{22.19} \\[0pt] \footnotesize\textcolor{green!40!black}{\textbf{(+1.74)}}} &
\shortstack[c]{\textbf{0.65} \\[0pt] \footnotesize\textcolor{green!40!black}{\textbf{(+0.03)}}} &
\shortstack[c]{\textbf{0.42} \\[0pt] \footnotesize\textcolor{green!40!black}{\textbf{(+0.02)}}} &
\shortstack[c]{\textbf{0.54} \\[0pt] \footnotesize\textcolor{green!40!black}{\textbf{(+0.02)}}} &
\shortstack[c]{\textbf{7.74} \\[0pt] \footnotesize\textcolor{green!40!black}{\textbf{(-2.97)}}} \\
\bottomrule
\end{tabular}
\label{tab:performance_feta}
\end{table}

\subsection{Main Results}

\subsubsection{Overall Performance}

Here, we compare CoQ with both generic and SQL-aided table understanding methods on three datasets involving multi-hop reasoning. As shown in Tables~\ref{tab:performance_wiki_fact} and \ref{tab:performance_feta}, CoQ consistently achieves both higher answer accuracy and lower SQL error rates. On WikiTQ, it reaches 74.77\% with GPT-3.5, a +13.66\% gain over the second-best. On TabFact, it attains 92.31\%, exceeding the best baseline by +10.39\%. On FeTaQA, CoQ reports the highest BLEU (22.19) and ROUGE scores (R-1: 0.65, R-2: 0.42, R-L: 0.54), and the lowest error rate (7.74\%). We also assess how CoQ scales with stronger models in Appendix~\ref{sec:appendix_more_experiments_strong_llms} and further analyze a challenging subset of WikiTQ in Appendix~\ref{sec:appendix_more_experiments_error}. These results highlight the effectiveness of CoQ, which is analyzed below.

\begin{table}[H]
\caption{Results of CoQ and hybrid baselines on WikiTQ and TabFact datasets (GPT-3.5). CoQ achieves the highest accuracy, attributed to its schema abstraction and fine-grained SQL generation control.}
\centering
\begin{tabular}{lcc}
\toprule
\textbf{Method} & \textbf{WikiTQ} & \textbf{TabFact} \\
\midrule
TabSQLify & 64.7 & 79.5 \\
TableMaster & 68.2 & 83.7 \\
H-STAR & 69.6 & \underline{85.0} \\
SynTQA & \underline{70.4} & N/A \\
\hline
\\[-10pt]
\raisebox{0.9ex}[0pt]{\textbf{Chain-of-Query (Ours)} } &
\shortstack[c]{\textbf{74.8} \\[-1pt] \footnotesize\textcolor{green!40!black}{\textbf{(+4.4)}}} &
\shortstack[c]{\textbf{92.3} \\[-1pt] \footnotesize\textcolor{green!40!black}{\textbf{(+7.3)}}} \\
\bottomrule
\end{tabular}
\label{tab:hybrid}
\end{table}

\begin{table*}[ht]
\caption{Results on IM-TQA and Open-WikiTable. Top-K = accuracy at top-K candidates. Note that Top-K accuracy is reported as a decimal rather than a percentage. CoQ maintains SOTA performance on structurally complex and multi-table datasets, enabled by its natural-language-style schema abstraction.}
\centering
\renewcommand{\arraystretch}{1}
\setlength{\tabcolsep}{6pt}
\begin{tabular}{lcc|cccc}
\toprule
\textbf{Method} &
\multicolumn{2}{c|}{\textbf{IM-TQA}} &
\multicolumn{4}{c}{\textbf{Open-WikiTable}} \\
\cmidrule(lr){2-3} \cmidrule(lr){4-7}
& Acc. & Inv. & Top-1 & Top-2 & Top-5 & Top-10 \\
\midrule
Few-Shot QA        & 52.47 & N/A   & 0.336 & 0.357 & 0.383 & 0.388 \\
Chain-of-Table     & 48.80 & N/A   & 0.404 & 0.430 & 0.463 & 0.467 \\
Basic Text-to-SQL  & 63.16 & 7.86  & 0.429 & 0.458 & 0.490 & 0.496 \\
OpenTab            & 67.28 & 6.94  & \underline{0.491} & \underline{0.523} & \underline{0.556} & \underline{0.565} \\
MAG-SQL            & \underline{68.90} & \underline{6.06}  & 0.457 & 0.476 & 0.516 & 0.524 \\
\midrule
\textsc{Chain-of-Query} (Ours) & 
\shortstack[c]{\textbf{74.96}\\[-2pt] \footnotesize\textcolor{green!40!black}{\textbf{(+6.06)}}} &
\shortstack[c]{\textbf{2.80}\\[-2pt] \footnotesize\textcolor{green!40!black}{\textbf{(-3.26)}}} &
\shortstack[c]{\textbf{0.527}\\[-2pt] \footnotesize\textcolor{green!40!black}{\textbf{(+0.036)}}} &
\shortstack[c]{\textbf{0.552}\\[-2pt] \footnotesize\textcolor{green!40!black}{\textbf{(+0.029)}}} &
\shortstack[c]{\textbf{0.592}\\[-2pt] \footnotesize\textcolor{green!40!black}{\textbf{(+0.076)}}} &
\shortstack[c]{\textbf{0.608}\\[-2pt] \footnotesize\textcolor{green!40!black}{\textbf{(+0.083)}}} \\
\bottomrule
\end{tabular}
\label{tab:imtqa_openwiki}
\end{table*}

\subsubsection{Comparison Against Generic Methods}

We now provide a detailed comparison with generic LLM-based table understanding methods, which rely solely on LLMs’ natural language capabilities without explicit SQL assistance. Their improvements are modest (e.g., Tree-of-Table improves accuracy by just 1.17\% over Chain-of-Table on WikiTQ). This indicates a performance plateau for current methods. On the other hand, Table-to-Text methods linearize entire tables into textual descriptions to eliminate structured content. However, the generated text often exhaustively covers all table content, making it difficult for LLMs to locate key information. This leads to low accuracy and poor scalability on large tables. Empirically, the table-to-text baseline performs the worst among all methods (e.g., 16.07\% on WikiTQ).

In contrast, CoQ consistently achieves substantial improvements across all datasets and LLM backbones. For example, it outperforms Tree-of-Table on WikiTQ by +13.66\% with GPT-3.5 and +14.90\% with LLaMA 2. Similar trends hold for TabFact and FeTaQA. These results confirm the effectiveness of CoQ’s SQL-aided design in enhancing the accuracy and robustness of structured reasoning. Additionally, its natural-language-style table schemas abstract away structural noise to avoid forcing LLMs to interpret tabular structures or irrelevant content.

\subsubsection{Comparison Against SQL Methods} 

We also compare CoQ with SQL-aided baselines and find that it consistently achieves higher accuracy and lower SQL error rates across all datasets and model backbones. A key limitation of existing SQL-aided methods lies in their rigid, one-shot formulation: they attempt to generate a single SQL query that directly yields the final answer. This approach works reasonably well for traditional Text-to-SQL tasks, where answers typically correspond to certain cell values. However, in table understanding, answers often require multi-step reasoning, abstraction, or comparisons across multiple rows or columns. Forcing all logic into one complex SQL query in such settings leads to bloated and error-prone programs. For example, while MAG-SQL performs well on standard Text-to-SQL benchmarks \citep{xie2024mag}, it underperforms on reasoning-intensive table understanding tasks. MAG-SQL only achieves 55.87\% accuracy and a 9.48\% invalid SQL rate on WikiTQ with GPT-3.5, compared to 74.77\% accuracy and just 3.34\% invalid SQL with CoQ. Similar trends are observed on TabFact and FeTaQA.

In contrast, CoQ benefits from our Clause-by-Clause SQL Generation, which ensures that generated queries remain concise and valid. Our Hybrid Reasoning Division further ensures that queries are simplified but aligned with the reasoning scope by offloading higher-level inference to the LLM. Moreover, our natural-language-style table schemas abstract away structural noise, reducing the LLM’s burden in interpreting complex layouts for SQL generation. As a result, CoQ achieves improved reliability without compromising reasoning depth.

\subsection{Comparison Against Hybrid Methods}

Here, we compare the performance of CoQ with hybrid methods on the WikiTQ and TabFact datasets using GPT-3.5. As shown in Table~\ref{tab:hybrid}, CoQ consistently achieves the best performance across both datasets.

Notably, although our work shares the same hybrid philosophy, it differs significantly in granularity and robustness. Unlike these approaches, which rely on simple SQL operations and one-shot SQL generation without invalid-query correction, CoQ employs a clause-by-clause validated SQL generation process with sufficiency checks and rollback mechanisms. This ensures that the generated SQL is both valid and adaptively complex for deeper mechanical reasoning. Logical inference is then handled by the LLM over the intermediate SQL outputs. This design achieves a more fine-grained separation between mechanical and logical reasoning while ensuring robustness. The key to this robust, fine-grained separation lies in dynamic planning, where the Planner adaptively adjusts the SQL generation depth depending on the input complexity.

\subsection{Generalization to Real-World Tables}
The previously explored datasets primarily feature clean, well-structured tables. However, real-world scenarios often involve messy, irregular tables. To evaluate CoQ's performance in such settings, we conduct additional experiments on the IM-TQA and Open-WikiTable datasets. IM-TQA contains structurally diverse (transposed, nested, and irregular formats) tables, while Open-WikiTable reflects multi-table databases in practical applications.

As shown in Table~\ref{tab:imtqa_openwiki}, CoQ maintains strong performance across both structurally complex and multi-table settings, consistent with our earlier results. A key contributor to this robustness is our natural-language-style schema, which abstracts away structural noise and allows LLMs to focus on semantic content rather than layout-specific details.

\begin{table}[ht]
\caption{Ablation results on WikiTQ (GPT-3.5), evaluating each CoQ component. CoQ achieves optimal performance through its modular design, with each component contributing significantly to accuracy and SQL validity.}
\centering
\renewcommand{\arraystretch}{0.85}
\setlength{\tabcolsep}{2pt}
\begin{tabular}{p{4.0cm} @{\hskip 8pt} l @{\hskip 17pt} l}
\toprule
\textbf{Method} & \textbf{Acc.} & \textbf{Inv.} \\
\midrule
\textsc{Chain-of-Query} & 
\makecell[c]{\textbf{74.77}} & 
\makecell[c]{\textbf{3.34}} \\[2pt]
\makecell[l]{w/o Natural-Language-\\[-4pt] 
 Schema} & 
\makecell[c]{64.65\\[-4pt] {\footnotesize\textcolor{red}{(-10.12)}}} & 
\makecell[c]{4.47\\[-4pt] {\footnotesize\textcolor{red}{(+1.13)}}} \\[4pt]
\makecell[l]{w/o Parallel Task\\[-4pt] 
 Decomposition} & 
\makecell[c]{72.95\\[-4pt] {\footnotesize\textcolor{red}{(-1.82)}}} & 
\makecell[c]{3.86\\[-4pt] {\footnotesize\textcolor{red}{(+0.52)}}} \\[4pt]
\makecell[l]{w/o Clause-by-Clause\\[-4pt] SQL Generation} & 
\makecell[c]{57.73\\[-4pt] {\footnotesize\textcolor{red}{(-17.04)}}} & 
\makecell[c]{8.82\\[-4pt] {\footnotesize\textcolor{red}{(+5.48)}}} \\[4pt]
\makecell[l]{w/o Hybrid Reasoning\\[-4pt] Division} & 
\makecell[c]{55.96\\[-4pt] {\footnotesize\textcolor{red}{(-18.81)}}} & 
\makecell[c]{5.27\\[-4pt] {\footnotesize\textcolor{red}{(+1.93)}}} \\
\bottomrule
\end{tabular}
\label{tab:ablation_wiki}
\end{table}

\subsection{Ablation Study: Deep Dive into CoQ’s Key Components and Mechanisms}

We conduct an ablation study on WikiTQ using GPT-3.5, disabling one CoQ component at a time. A more detailed analysis is provided in Appendix~\ref{sec:appendix_more_experiments_ablation}.

As shown in Table~\ref{tab:ablation_wiki}, removing natural-language-style schemas leads to a clear performance drop (–10.12\%) and more invalid SQL (+1.13\%). This highlights the benefit of abstracting away structural noise. Removing the Parallel Task Decomposition causes a minor accuracy drop (–1.82\%), suggesting its limited impact on simple cases but usefulness in complex ones. Disabling the Clause-by-Clause SQL Generation causes a substantial drop in accuracy (–17.04\%) and a sharp rise in invalid SQL (+5.48\%), as clause-level generation supports incremental validation and accurate clause selection. This reduces the risk of incorrect queries and ensures appropriate SQL operations are applied. The largest drop (–18.81\%) occurs without the Hybrid Reasoning Division, which prevents over-generation and delegates logical reasoning to the LLM. Without this control, the system tends to over-generate, increasing query complexity without improving answer quality.

Overall, the ablation results confirm that each CoQ component contributes meaningfully to performance, highlighting the effectiveness of its modular design in balancing precision and adaptability.

\begin{table}
\caption{Comparison of the theoretical number of LLM calls per question across methods on WikiTQ dataset.}
\centering
\renewcommand{\arraystretch}{1}
\setlength{\tabcolsep}{12pt}
\begin{tabular}{l r}  
\toprule
\textbf{Method} & \textbf{\# Calls} \\
\midrule
Binder & 50 \\
Dater & 100 \\
Chain-of-Table & $\leq$ 25 \\
Tree-of-Table & $\leq$ 29 \\
MAC-SQL & $\leq$ \textbf{19} \\
MAG-SQL & $\leq$ 25 \\
\textsc{Chain-of-Query} & $\leq$ \underline{22} \\
\bottomrule
\end{tabular}
\label{tab:samples_wiki}
\end{table}

\subsection{Analysis of LLM Usage}
In addition to accuracy, we examine CoQ's LLM usage by analyzing the theoretical upper bound of calls required per question (Table~\ref{tab:samples_wiki}). Among all methods, CoQ offers a favorable trade-off between reasoning depth and LLM usage, requiring at most 22 calls. In contrast, other strong baselines require up to 25–100 calls due to extensive branching and fixed steps. While MAC-SQL is slightly more efficient (within 19 calls), it underperforms on complex tasks. Notably, we report upper bounds here; in practice, \textbf{CoQ averages just 7.63 LLM calls per instance} (Appendix~\ref{sec:appendix_empirical_calls}). This low average stems from our Hybrid Reasoning Division: by delegating logical reasoning to the LLM, CoQ avoids generating overly complex SQL and reduces error correction. It issues only the essential clauses needed to retrieve sufficient information, minimizing execution overhead. A detailed step-wise breakdown of LLM calls and an analysis of token-level cost are provided in Appendix~\ref{sec:appendix_more_experiments_cost}.

\section{Conclusion}

We introduce \textsc{Chain-of-Query}, a multi-agent framework for SQL-aided table understanding, built upon three key insights: natural-language-style schema, clause-by-clause generation and hybrid reasoning division. They enable robust, interpretable, and precise reasoning over tables. CoQ achieves SOTA performance across five benchmarks, demonstrating strong generalization with low error rates and few LLM calls. It offers insights into effective integration of reasoning with structured query assistance in complex tabular tasks.

\section*{Limitations}

While our experiments span five benchmarks across diverse domains in both English and Chinese, and \textsc{Chain-of-Query} demonstrates strong generalization across datasets within these two languages, its effectiveness in other languages remains unexplored. Future work could extend the framework to multilingual settings and evaluate its adaptability to a wider range of linguistic phenomena.

\section*{Ethics Statement}

This work uses five publicly available datasets: WikiTQ, TabFact, FeTaQA, IM-TQA, and Open-WikiTable, all of which are widely used in prior research and contain no personally identifiable information. No additional data collection or human annotation was performed. Our experiments use GPT-3.5, GPT-4.1, LLaMA 2, LLaMA 3.1, and DeepSeek-V3 in standard inference-only settings. While we are not aware of specific ethical concerns related to our datasets or methods, we acknowledge potential risks: the reliance on pretrained LLMs may propagate social or cultural biases, and large-scale model deployment could raise environmental concerns.

\section*{Acknowledgments}

We thank all of our team members for their dedicated efforts and valuable discussions throughout this project, especially during the author–reviewer discussion phase. We thank the authors of Chain-of-Table, MAG-SQL, and OpenTab for sharing their code. Lastly, we sincerely appreciate all reviewers for their invaluable feedback, insightful suggestions, and positive remarks about our work.

\bibliography{refs} 

\clearpage 

\appendix

\section{Case Study}
\label{sec:appendix_case}

\subsection{Hybrid Reasoning}

\begin{table}[ht]
\centering
\small 
\caption{Case study: Fabrice Santoro's Grand Slam results and win--loss record.}
\renewcommand{\arraystretch}{1.05}
\setlength{\tabcolsep}{5pt} 
\begin{tabular}{lccc}
\toprule
\textbf{Name} & \textbf{2001} & \textbf{2002} & \textbf{n\_win\_loss} \\
\midrule
Australian Open & 2R & 3R & 22--18 \\
French Open     & 2R & 2R & 17--20 \\
Wimbledon       & 2R & 1R & 11--14 \\
\bottomrule
\end{tabular}
\label{tab:santoro_case}
\end{table}

\noindent\textbf{Question: Did Fabrice Santoro win more at the Australian Open or Wimbledon?}

\begin{figure}[htbp]
  \centering
  \includegraphics[width=\linewidth]{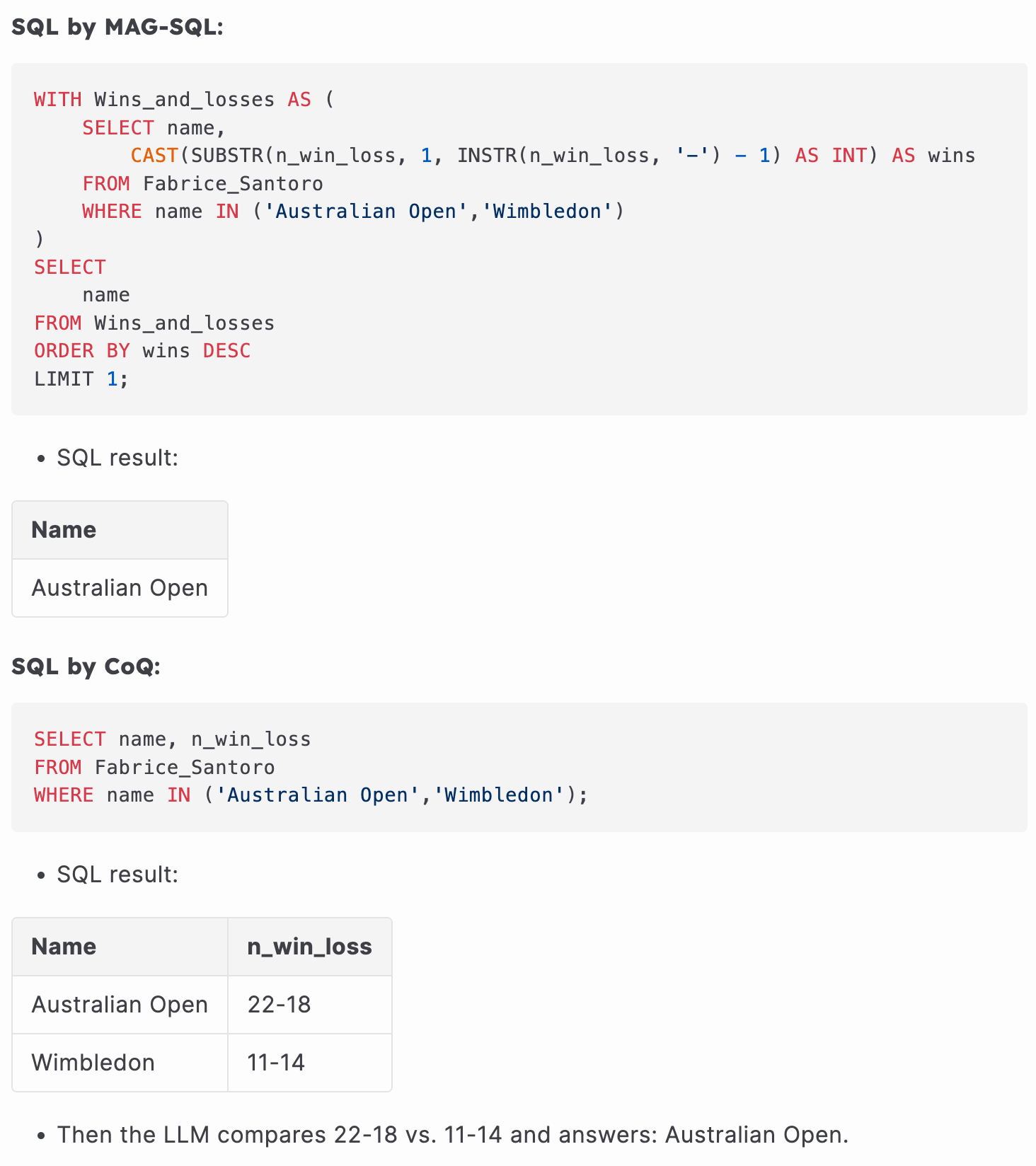}
  \caption{Comparison of MAG-SQL vs. \textsc{Chain-of-Query}.}
  \label{fig:case_compare}
\end{figure}

This example highlights the fundamental difference between MAG-SQL and \textsc{Chain-of-Query} in handling structured reasoning.

As shown in Figure~\ref{fig:case_compare}, MAG-SQL attempts to answer the question entirely within SQL. It constructs a nested query that parses the \texttt{n\_win\_loss} field (e.g., 22-18) to extract the number of wins using string manipulation functions like \texttt{SUBSTR} and \texttt{INSTR}. It then casts the result to integer and selects the entry with the highest number of wins. While functional, this query is long, brittle, and tightly coupled to a specific string format. Any minor format mismatching would cause execution failure or incorrect results. This reflects a common issue with traditional SQL-aided approaches: in pursuit of returning the final answer in one shot, they overcomplicate query logic and increase error risk.

CoQ, in contrast, adopts a different strategy. It first generates a simple SQL query that merely retrieves the relevant rows and raw \texttt{n\_win\_loss} records. Then, the LLM compares the values ("22–18" vs. "11–14") and reasons that the Australian Open had more wins. This hybrid division—SQL for extraction, LLM for comparison—reduces SQL complexity and improves robustness, as the LLM is better suited to handle slight variations or contextual interpretations in text-formatted values.

In short, MAG-SQL complicates SQL to achieve end-to-end reasoning, whereas CoQ simplifies SQL and leverages the LLM where it excels, demonstrating the strength of our Hybrid Reasoning Division Strategy.

\subsection{Handling String Variations}

For noisy entries such as "New York" vs. "new york\texttt{*}" or "Dallas Cowboys" vs. "Dallas Cowboys\texttt{*}", our framework addresses the issue through fuzzy matching in SQL generation and refinement.

For example, in questions such as "How many games did the cowboys play?", the LLM generates SQL queries like \texttt{SELECT ... WHERE team LIKE "\%Dallas Cowboys\%"}.

Appendix~\ref{appendix_where} shows our prompt for correcting invalid WHERE clauses, and its "Constraints" section encourages the LLM to relax filtering conditions, including using fuzzy matching "IN" and "LIKE".

This mechanism ensures that variations with minor inconsistencies are still matched. While fuzzy matching cannot cover all possible noise patterns, in practice it significantly reduces such errors and allows CoQ to remain robust to imperfect table entries.

\subsection{Handling Special Rows}

The key challenge of handling such rows lies in their special values. In our framework, these rows are addressed through the integration of the Natural-Language-Style Table Schema and Hybrid Reasoning.

The schema provides representative value examples for each column, which are typically normal values randomly sampled from the table. Our SQL-based mechanical reasoning does not directly yield the final answer but instead produces an intermediate sub-table. This allows the LLM to compare the schema examples with the SQL results and detect anomalies such as special values from the aggregate rows.

\begin{figure}[t]
  \centering
  \includegraphics[width=\linewidth]{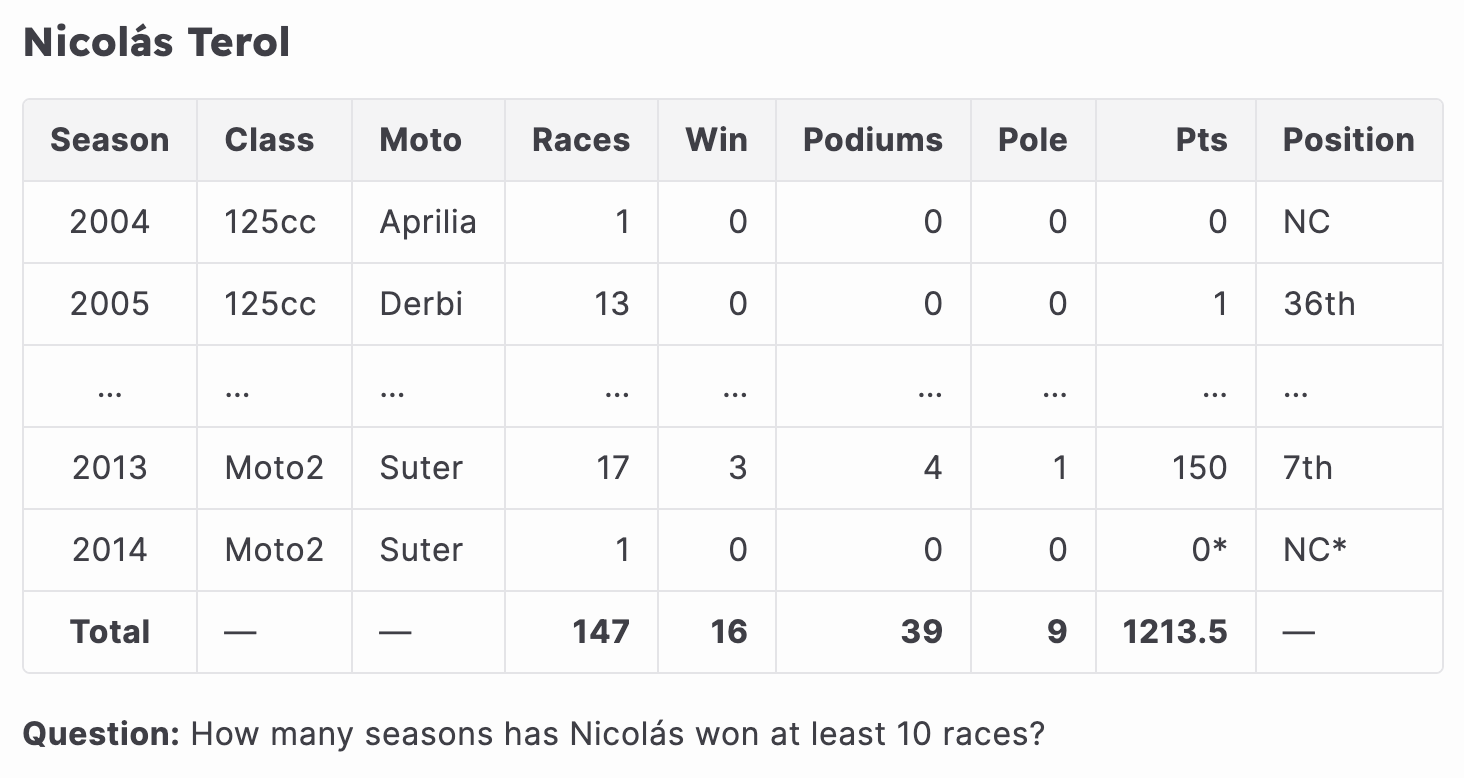}
  \caption{An example of a table containing a special row, where the last row includes "Total".}
  \label{fig:special_row}
\end{figure}

\begin{figure}[htbp]
  \centering
  \includegraphics[width=\linewidth]{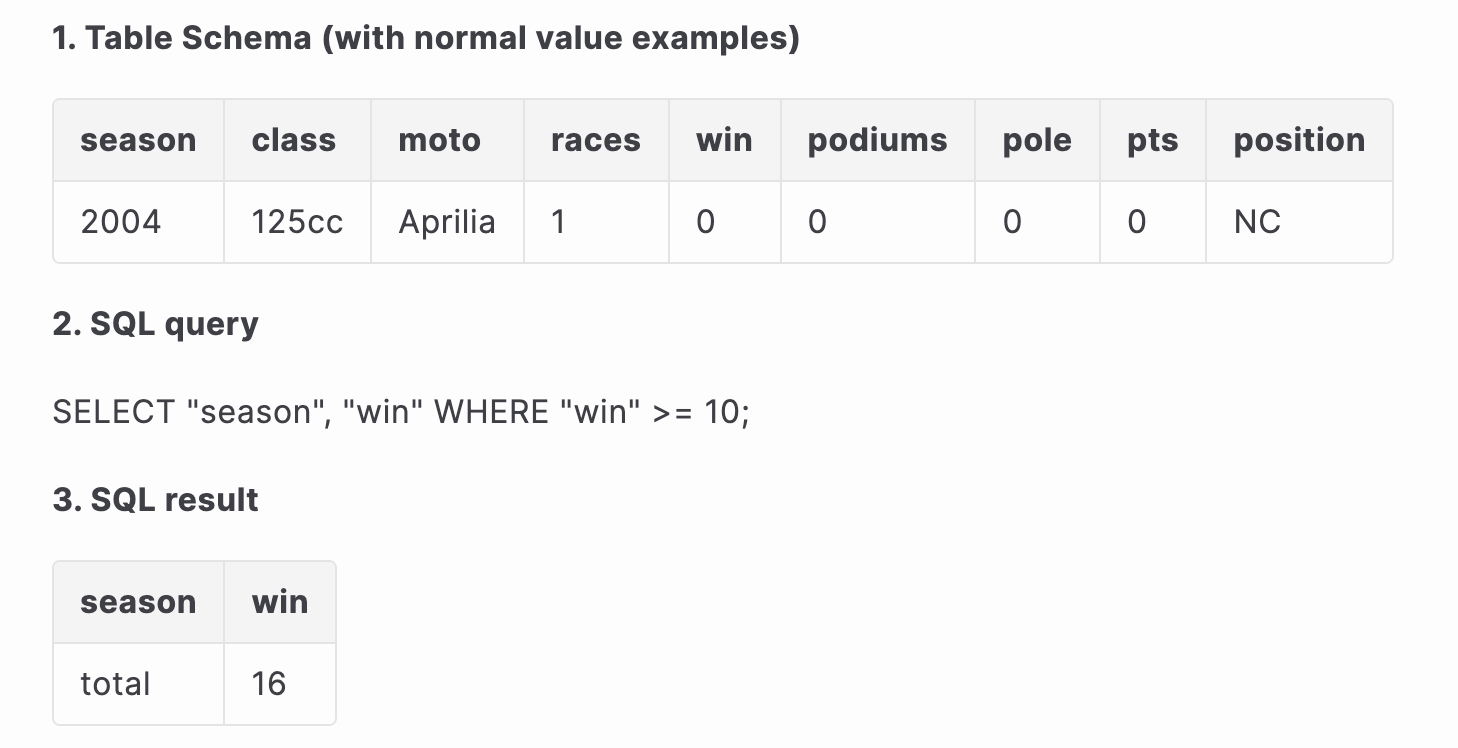}
  \caption{Illustration of CoQ’s intermediate information in this case.}
  \label{fig:coq_row}
\end{figure}

For instance, Figure~\ref{fig:special_row} shows a table with special rows studied by \citet{liu2023rethinkingtabulardataunderstanding}.In their framework, the Python-based method attempts to directly return the exact number, counting the aggregate row by mistake. However, our method provides the LLM with intermediate information as illustrated in Figure~\ref{fig:coq_row}.

In this example, the LLM can distinguish the normal entry "2004" from the special entry "total" and correctly infer that the result contains an aggregate row, thus avoiding misinterpretation.

\section{Examples of \textsc{Chain-of-Query}}

\subsection{SQL Generation Example Using Parallel Decomposition}
\label{sec:appendix_decompose_example}

\begin{figure}[htbp]
  \centering
  \includegraphics[width=\linewidth]{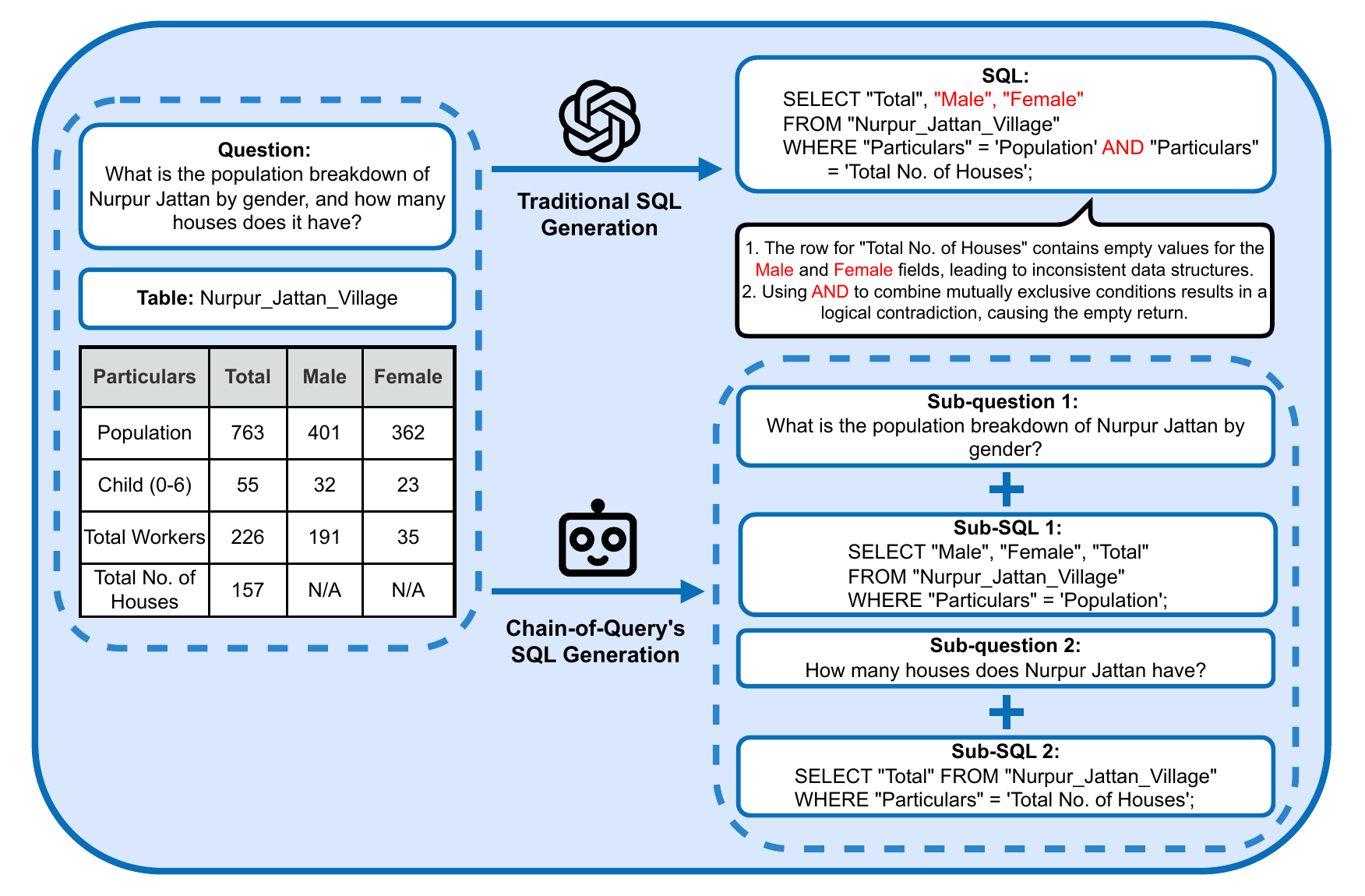}
  \caption{Comparison of traditional vs. \textsc{Chain-of-Query} SQL generation via parallel sub-question decomposition.}
  \label{fig:decomposer_compare}
\end{figure}

Figure~\ref{fig:decomposer_compare} illustrates the difference between traditional SQL generation and our \textsc{Chain-of-Query} framework using parallel sub-question decomposition. In the traditional approach, the LLM is required to answer the entire question using a single SQL query, which leads to structurally complex queries. Due to limited understanding of both the table structure and SQL semantics, the model may incorrectly use conjunctions like AND to combine incompatible conditions. This often results in logical inconsistencies or empty outputs, especially when the involved rows contain mismatched structures or missing values.

In contrast, \textsc{Chain-of-Query} decomposes the original question into two independent sub-questions, each focusing on a distinct aspect of the table. These sub-questions are processed in parallel and translated into simpler, more focused SQL queries. This strategy improves robustness, avoids conflicting constraints, and simplifies table access by localizing reasoning to narrower regions.

\subsection{SQL Generation Example with Early Stopping}
\label{sec:appendix_planner_example}

\begin{figure}[htbp]
  \centering
  \includegraphics[width=\linewidth]{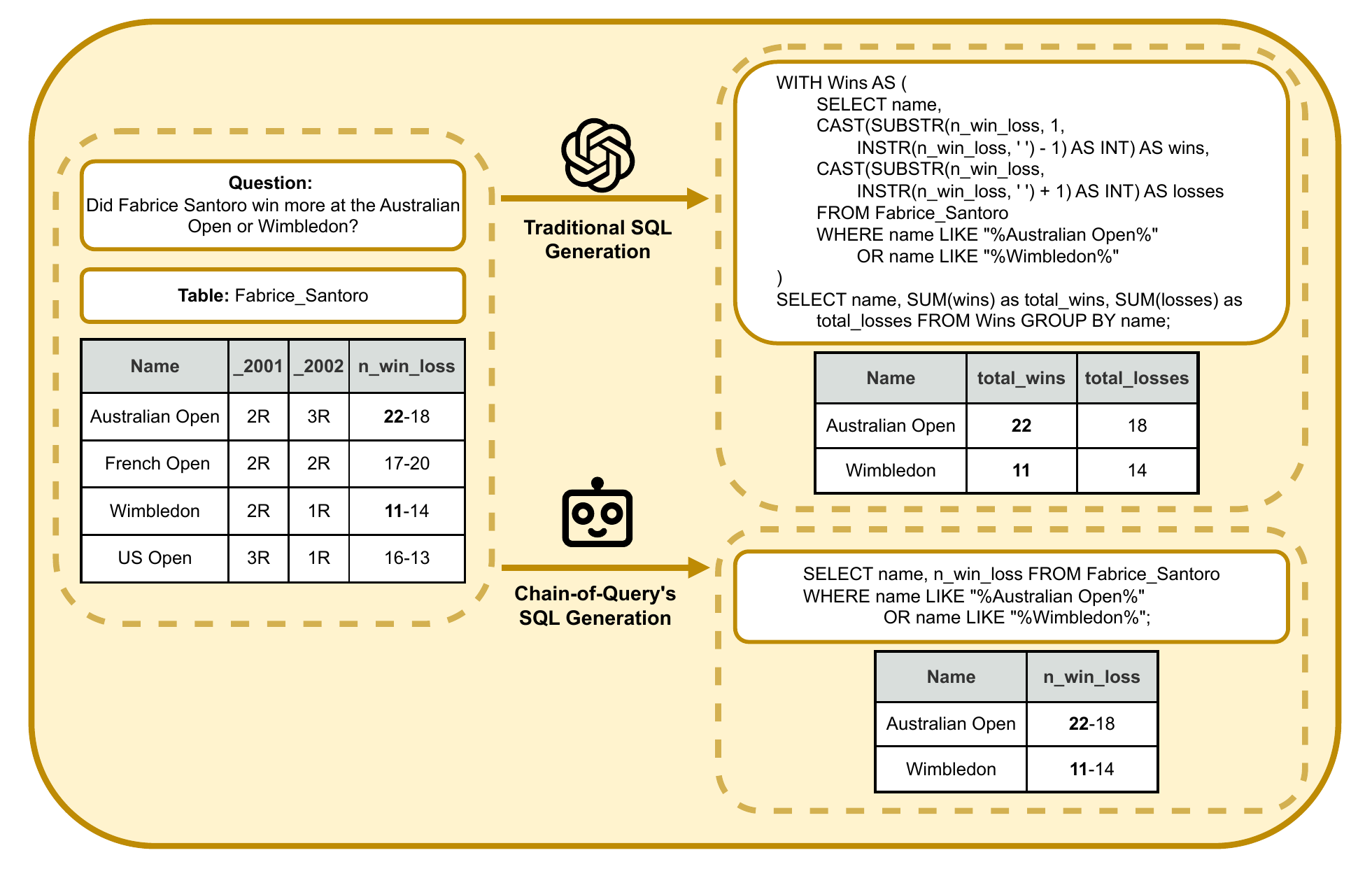}
  \caption{Example of Sufficiency-based Early Stopping in SQL generation.}
  \label{fig:planner_compare}
\end{figure}

Figure~\ref{fig:planner_compare} illustrates the difference between traditional SQL generation and our \textsc{Chain-of-Query} framework with the Sufficiency-based Early Stopping Mechanism. While the traditional approach constructs a complex query involving nested operations (e.g., \texttt{WITH}, \texttt{CAST}, \texttt{SUBSTR}) to compute explicit win/loss counts, our method halts early once sufficient information has been retrieved. Specifically, our method issues a simpler SQL query that directly extracts raw win–loss strings (e.g., "22–18", "11–14") from the relevant rows. The LLM then completes the comparison based on these outputs. This early stopping mechanism reduces unnecessary query complexity and enhances robustness by delegating the final reasoning step to the language model. Notably, despite its increased complexity, the traditional query yields essentially the same informational content as the simpler query used in our approach. Moreover, such complexity introduces more opportunities for execution failures and semantic mismatches.

\section{Clause Option List of Clause-by-Clause SQL Generation Strategy}
\label{sec:appendix_clause}

Based on common patterns of information retrieval and reasoning operations in table understanding tasks, we define five types of SQL clauses used in our clause-by-clause generation strategy:

\begin{itemize}
    \item \texttt{SELECT-FROM} clause: Serves as the foundational component of an SQL query, used to select task-relevant columns from the table.
    
    \item \texttt{WHERE} clause: Filters table rows based on specific conditions relevant to the question.
    
    \item \texttt{WITH AS} clause: Defines a Common Table Expression (CTE), enabling the creation of temporary virtual tables for intermediate transformations. This includes generating new columns or modifying existing ones without altering the original table. Such operations help organize and extract new information to support downstream reasoning.
    
    \item Aggregate function clause: Applies aggregation functions such as \texttt{COUNT}, \texttt{SUM}, \texttt{AVG}, \texttt{MAX}, and \texttt{MIN} to summarize or compute over table content. These operations are often essential for high-level reasoning and abstraction.

    \item \texttt{ORDER BY} clause: Sorts the table based on specified columns, facilitating reasoning that depends on ranking or positional relationships in the data.
\end{itemize}

\section{Algorithm of Sufficiency-based Early Stopping Mechanism}
\label{sec:appendix_stopping}

We present the pseudocode for the Sufficiency-Based Early Stopping Mechanism, which allows the Dynamic Planner to terminate clause generation once sufficient information has been retrieved.

\renewcommand{\AlCapFnt}{\small}
\begin{algorithm}
\small
\caption{{\small\textsc{Sufficiency-based Early Stopping}}}
\KwData{\scalebox{0.9}{ $(Q, T)$, where Q is a natural language question;}\\
\scalebox{0.9}{\hspace*{3.5em}$T = (S, D)$ is a  table consisting of schema $S$ and}\\
\scalebox{0.9}{\hspace*{3.5em}data content $D$.}}
\KwResult{\scalebox{0.9}{$\hat D$ is the extracted subset of table's data content}\\
\scalebox{0.9}{\hspace*{3.9em}used to answer the question.}}
\DontPrintSemicolon
\Fn{\texttt{Sufficiency-Early-Stop ($Q, T$)}:}{
  $\mathrm{chain} \gets \texttt{[Generator(Q, T)]}$\;
  \Repeat{$\mathrm{plan} = \text{STOP}$}{
    $\mathrm{sql} \gets \mathrm{chain}[-1]$\;
    $D' \gets \texttt{ExecuteSQL(T, sql)}$\;
    $\mathrm{plan} \gets \texttt{Planner(Q, S, sql, D')}$\;
    \If{$\mathrm{plan} \neq \text{STOP}$}{
      $\mathrm{next\_sql} \gets \texttt{Generator(Q, T, chain, plan)}$\;
      $\mathrm{chain}.\mathrm{append}(\mathrm{next\_sql})$\;
    }
  }
  $\hat D \gets D'$\;
  \Return{$\hat D$}\;
}
\end{algorithm}

\section{Empirical Statistics on LLM Calls per Instance}
\label{sec:appendix_empirical_calls}

\begin{figure}[htbp]
  \centering
  \includegraphics[width=\linewidth]{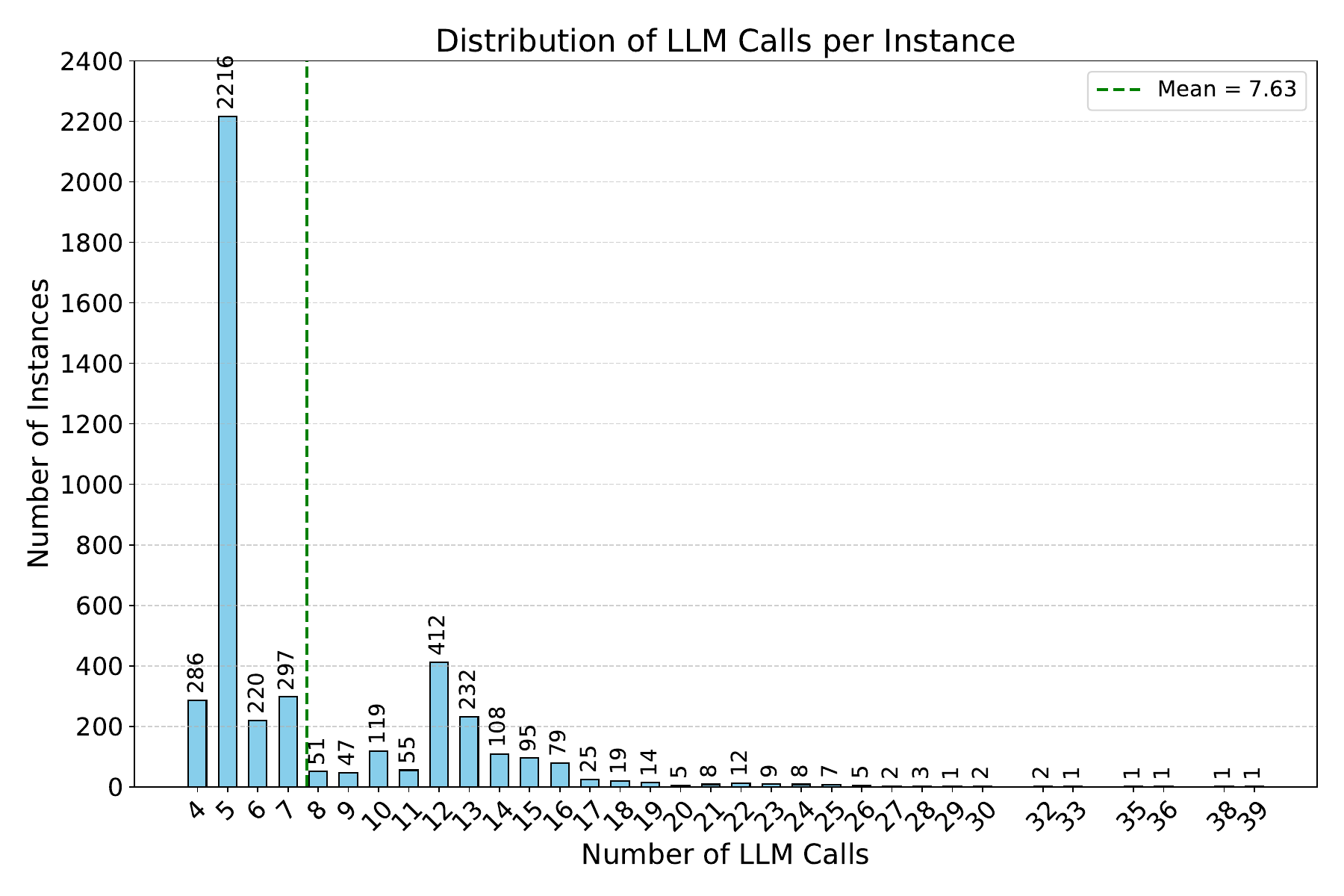}
  \caption{Empirical LLM Calls per Instance (WikiTQ). CoQ completes most instances within very few LLM calls, enabled by hybrid reasoning, early stopping, and clause-by-clause generation.}
  \label{fig:llm_calls}
\end{figure}

For WikiTQ, as shown in Figure~\ref{fig:llm_calls}, CoQ requires \textbf{on average only 7.63 LLM calls per instance}, with a median of just 5. Notably, 57\% of the instances complete within 5 calls, and only 0.16\% exceed 30 calls, with the maximum observed at 39. These statistics highlight the low cost of our approach, especially when compared to other multi-agent baselines that incur significantly higher LLM usage.

This low usage arises from two key factors. First, our Hybrid Reasoning Division and Early Stopping prevent the construction of overly long SQL queries by stopping clause generation once sufficient information has been retrieved, avoiding unnecessary or overly complex SQL clauses. Second, our Clause-by-Clause Generation allows the LLM to focus on one clause at a time rather than constructing an entire SQL query at once. This not only simplifies generation at each step but also substantially reduces the chance of syntax or logic errors, minimizing the need for repeated LLM corrections. Together, these strategies ensure that our method maintains high SQL generation quality and minimizes redundant LLM usage.

\section{Dataset Details}
\label{sec:appendix_data}

We use five publicly available datasets in our experiments: \textbf{WikiTQ}, \textbf{TabFact}, \textbf{FeTaQA}, \textbf{IM-TQA}, and \textbf{Open-WikiTable}. Below we summarize their features, sources, formats, licensing, and usage details.

\subsection{WikiTQ}
WikiTQ \citep{pasupat2015wikitab} is a question answering dataset derived from Wikipedia, a resource with broad topical coverage. It contains 2,108 tables and 22,033 natural language questions, each paired with an answer derivable from a single table. Answers typically correspond to one or more table cells.

\noindent\textbf{License}: CC BY-SA.\\
\textbf{Language}: English.\\
\textbf{Split Used}: Standard test set (2,273 examples) following prior work.\\
\textbf{Content}: No personally identifiable or offensive content observed.

\subsection{TabFact}
TabFact \citep{2019TabFactA} is a table-based fact verification dataset over Wikipedia tables covering over 16 domains. Each example includes a table and a natural language statement, labeled as either entailed or refuted. It comprises 117,854 examples with binary labels.

\noindent\textbf{License}: CC BY-SA.\\
\textbf{Language}: English.\\
\textbf{Split Used}: Standard test set (12,779 examples) following prior work.\\
\textbf{Content}: No personally identifiable or offensive content observed.

\subsection{FeTaQA}
FeTaQA \citep{Nan2021FeTaQAFT} is a complex table QA dataset focused on multi-hop reasoning. Each entry includes a Wikipedia table, a natural language question, and a free-form textual answer. It focuses on free-form, multi-step reasoning beyond cell-level lookup.

\noindent\textbf{License}: CC BY-SA.\\
\textbf{Language}: English.\\
\textbf{Split Used}: Standard test set (2,003 examples) following prior work.\\
\textbf{Content}: No personally identifiable or offensive content observed.

\subsection{IM-TQA}
IM-TQA \citep{zheng-etal-2023-im} is a table QA dataset that is built by collecting tables from open websites of more than 10 domains. It features complex table styles, including hierarchical, nested, and messy real-world structures. Each entry includes a structurally complex table, several natural language questions, and answers relevant to certain cells. 

\noindent\textbf{License}: CC BY-SA.\\
\textbf{Language}: Chinese.\\
\textbf{Split Used}: Standard test set (464 examples) following prior work.\\
\textbf{Content}: No personally identifiable or offensive content observed.

\subsection{Open-WikiTable}
Open-WikiTable \citep{kweon2023openwikitabledatasetopendomain} is a dataset for open domain question answering with complex reasoning over multi-table settings. Each entry includes a natural language question, a table related to the question, and the answer.

\noindent\textbf{License}: CC BY-SA.\\
\textbf{Language}: English.\\
\textbf{Split Used}: Standard test set (6,602 examples) following prior work.\\
\textbf{Content}: No personally identifiable or offensive content observed.

\subsection{Usage and Compliance}
We use all datasets strictly for academic research in inference-only mode, without additional annotation or redistribution. Usage is fully aligned with the datasets’ intended purposes and license terms.

\section{Reproducibility Statement}
\label{sec:appendix_reproduce}

\subsection{\textsc{Chain-of-Query}}

We conduct all experiments using the following models as the backbone LLMs: GPT-3.5-turbo \citep{brown2020languagemodelsfewshotlearners}, GPT-4.1 \citep{openai2024gpt4technicalreport}, LLaMA-2-13B \citep{touvron2023llama2openfoundation}, LLaMA-3.1-8B \citep{grattafiori2024llama3herdmodels}, and DeepSeek-V3 \citep{deepseekai2025deepseekv3technicalreport}. Model configurations are provided in Appendix~\ref{sec:appendix_parameters}. Prompt examples for each agent in the \textsc{Chain-of-Query} framework are provided in Appendix~\ref{sec:appendix_prompts_coq}.

We also employ a lightweight memory mechanism for the SQL Query Generator. Specifically, we maintain a list of validated SQL queries during clause-by-clause generation. Each newly generated clause is executed; if valid, it is appended to the list. If it fails and cannot be corrected, the system automatically falls back to the last validated query stored in memory. This ensures robustness and prevents error propagation without requiring a heavy memory module.

\subsection{Baselines}

Our implementation is based on the same environment as Chain-of-Table~\citep{chaintab_github}. To ensure fair comparison, we directly reuse the GPT-3.5 and LLaMA 2 results for the following baselines as reported in their paper: Few-Shot QA, Chain-of-Thought, Binder, and Dater. 

We re-run OpenTab, MAC-SQL, and MAG-SQL using their official open-source implementations~\citep{opentab_github, macsql_github, magsql_github}, following the same inference settings described in their respective repositories.

We used the Table-to-Text prompt provided by \citep{min2024exploringimpacttabletotextmethods} to obtain text format tables as one baseline.

Prompt examples for Basic Text-to-SQL (BT2SQL) and End-to-End QA (E2E QA) are included in Appendices \ref{sec:appendix_prompts_bt2sql} and \ref{sec:appendix_prompts_e2eqa}, respectively.

\subsection{Tooling and Package Settings}

All experiments are conducted in a Python 3.10 environment using standard open-source libraries. We make no modifications to any third-party packages beyond parameter configuration.

For preprocessing and database operations, we use \texttt{pandas} for basic table parsing and manipulation with no manual adjustments, and access SQLite databases via the \texttt{sqlite3} module along with the lightweight \texttt{records} wrapper.

We further evaluated CoQ on a temporal multi-table setting based on the Open-WikiTable dataset, following the setup of the baseline \citep{opentab_github}. This setup involves table retrieval from a large corpus using BM25 to select the top-k tables.

To evaluate natural language outputs, we report BLEU and ROUGE scores. BLEU is computed using the \texttt{nltk} library, while ROUGE is calculated using the \texttt{rouge-score} package, both with default settings. Accuracy is computed using the official evaluation scripts provided by the dataset authors.

Additionally, we use a custom script to compute the invalid SQL rate, defined as the proportion of model-generated SQL queries that fail to execute due to syntax or runtime errors.

\section{LLMs' Inference Configurations}
\label{sec:appendix_parameters}

To ensure the reproducibility and stability of LLM outputs across all agents in our framework, we carefully configure the decoding parameters for each model.

For \texttt{GPT-3.5-turbo} model, \texttt{GPT-4.1} model, and \texttt{DeepSeek-V3} model, we adopt a deterministic configuration: temperature is set to 0.0 and \texttt{top\_p} to 1.0. This disables sampling and enforces greedy decoding, ensuring consistent outputs across repeated runs. All responses are generated using official APIs from OpenAI and DeepSeek. Results are reported from a single run, without sampling variability.

For \texttt{LLaMA 2} and \texttt{LLaMA 3.1}, we deploy the 13B-chat and the 8B-instruct models respectively using Hugging Face Inference Endpoints, hosted on 8 A100 GPUs. Due to platform constraints, exact settings of \texttt{temperature} = 0.0 and \texttt{top\_p} = 1.0 are not available. As a practical approximation, we set \texttt{temperature} = 0.01 and \texttt{top\_p} = 0.9 for all agents, which yields nearly deterministic outputs while maintaining compatibility with the deployment platform. All LLaMA results are likewise reported from a single inference run.

\section{More Related Work}
\label{sec:appendix_more_related}

\begin{table*}[ht]
\caption{Results on WikiTQ with LLaMA~3.1, DeepSeek-V3, and GPT-4.1. 
CoQ delivers greater gains than upgrading to stronger LLMs, enabled by hybrid reasoning and robust SQL generation.}
\centering
\renewcommand{\arraystretch}{1}
\setlength{\tabcolsep}{4pt}
\begin{tabular}{lcccccccc}
\toprule
\multirow{2}{*}{\textbf{Method}} 
& \multicolumn{2}{c}{\textbf{LLaMA~3.1}} 
& \multicolumn{2}{c}{\textbf{DeepSeek-V3}} 
& \multicolumn{2}{c}{\textbf{GPT-4.1}} \\
\cmidrule(lr){2-3} \cmidrule(lr){4-5} \cmidrule(lr){6-7}
& Acc. & Inv. & Acc. & Inv. & Acc. & Inv. \\
\midrule
Table-to-Text   & 16.41 & N/A   & 48.39 & N/A   & 50.05 & N/A \\
Few-Shot QA     & 39.94 & N/A   & 66.07 & N/A   & 70.97 & N/A \\
Chain-of-Table  & \underline{54.17} & N/A 
                & \underline{71.64} & N/A 
                & \underline{75.78} & N/A \\
Basic Text-to-SQL & 44.13 & 16.71  
                & 64.75 & 12.10 
                & 70.79 & 7.50 \\
MAG-SQL         & 45.49 & \underline{11.44}  
                & 65.79 & \underline{9.14}  
                & 71.71 & \underline{5.82} \\
\midrule
\textsc{Chain-of-Query} (Ours) &
\shortstack[c]{\textbf{62.18}\\[-2pt] \footnotesize\textcolor{green!40!black}{\textbf{(+8.01)}}} &
\shortstack[c]{\textbf{5.62}\\[-2pt] \footnotesize\textcolor{green!40!black}{\textbf{(-5.82)}}} &
\shortstack[c]{\textbf{81.85}\\[-2pt] \footnotesize\textcolor{green!40!black}{\textbf{(+10.21)}}} &
\shortstack[c]{\textbf{2.30}\\[-2pt] \footnotesize\textcolor{green!40!black}{\textbf{(-6.84)}}} &
\shortstack[c]{\textbf{84.92}\\[-2pt] \footnotesize\textcolor{green!40!black}{\textbf{(+9.14)}}} &
\shortstack[c]{\textbf{1.54}\\[-2pt] \footnotesize\textcolor{green!40!black}{\textbf{(-4.28)}}} \\
\bottomrule
\end{tabular}
\label{tab:wikiq_sota_models}
\end{table*}

\subsection{Multi-Agent Table Understanding.} Recent work has explored LLM-powered agents \citep{wu2023autogen, Wang_2024agent} that perform multi-step reasoning and invoke table operation tools. Dater \citep{ye2023dater} reformulates questions into cloze-style prompts and retrieves values via LLM-generated SQL. Chain-of-Table \citep{wang2024chaintab} and Tree-of-Table \citep{ji2024treetab} construct dynamic reasoning paths using pre-defined Python functions. Although effective, these approaches rely on coding-based steps. This makes them fragile, as code errors can cause the entire reasoning chain to fail. For example, \citet{zhou2025efficientmultiagentcollaborationtool} reports that about half of the failures come from invalid code generation. Moreover, most of these methods are limited by fixed function sets, restricting their reasoning flexibility.

\subsection{General Structured Data Reasoning with LLMs}
Recent studies have extended LLM reasoning from plain text to structured data such as tables, knowledge graphs, and knowledge bases. Several general frameworks can also be adapted for table understanding. Pangu \citep{gu2023dontgeneratediscriminateproposal} generates multiple symbolic sub-expression candidates and lets an LLM select among them, but its semantic (rather than executable) evaluation causes error propagation once an incorrect candidate is retained. Chain-of-Knowledge \citep{li2024chainofknowledgegroundinglargelanguage} trains an adaptive query generator to produce symbolic queries that LLMs later validate, but its one-shot generation yields long and fragile expressions. StructGPT \citep{jiang2023structgptgeneralframeworklarge} reformulates prompts for column- and row-level filtering, avoiding invalid queries but treating tables as plain text, while Readi \citep{cheng2024necessaryllmsefficientlyfaithfully} generates shallow reasoning paths over key headers and entities, limiting support for complex manipulations. QueryAgent \citep{huang2024queryagentreliableefficientreasoning} supports SQL generation from the perspective of knowledge bases, yet lacks sufficient logical reasoning over tables.

While some high-level principles have been successfully applied in general structured data reasoning, especially in knowledge base question answering (KB-QA), our contribution lies in adapting and extending them into a modular, table-specific, multi-agent SQL framework, where these approaches have not been systematically investigated.

Our framework differs from KB-QA counterparts in fundamental ways: \textbf{1.} \emph{Question decomposition.} KB-QA methods typically decompose via relation/path traversal (hop-by-hop). Our work casts decomposition as semantic sub-questions and SQL sub-query evolution driven by LLM agents. \textbf{2.} \emph{Natural-language-style schema.} KB-QA methods often replace numerical KB IDs with surface names. In contrast, we construct a natural-language-style schema (table name, header–type pairs, and sampled example values) to represent tabular structures in natural language, eliminating interference from messy table layouts during LLM reasoning. \textbf{3.} \emph{Step-by-step correction.} KB-QA methods often apply post-hoc fixes to relation bindings. Our framework integrates clause-by-clause SQL generation with execution validation and targeted correction, ensuring robustness during query construction rather than patching afterward.

\subsection{Broader Applications of Large Language Models}
The application of deep learning and large language models beyond pure text data is not limited to table understanding. Broader domains such as recommendation systems \citep{10884167, 10.1145/3626772.3657807, qu2025tokenrec, zhang-etal-2025-cove} and time series analysis \citep{chuang2025ltsmbundletoolboxbenchmarklarge, zhou-etal-2025-merit, zhou2025enhancingllmreasoningtime, sui2025trainingfreetimeseriesclassification} are also important in practice. Extending and further validating our multi-agent hybrid reasoning strategy in these broader settings represents a promising direction for future work.

\begin{table}[H]
\caption{Performance comparison of CoQ and general structured data reasoning baselines on WikiTQ and TabFact datasets (GPT-3.5). CoQ achieves the highest accuracy, attributed to its hybrid reasoning and clause-wise SQL generation.}
\centering
\renewcommand{\arraystretch}{0.95}
\setlength{\tabcolsep}{6pt}
\begin{tabular}{lcc}
\toprule
\textbf{Method} & \textbf{WikiTQ} & \textbf{TabFact} \\
\midrule
QueryAgent & 52.8 & 59.1 \\
StructGPT & 52.2 & \underline{87.6} \\
Readi & 61.7 & N/A \\
Chain-of-Knowledge & \underline{70.4} & N/A \\[2pt]

\hline
\\[-10pt]
\raisebox{0.9ex}[0pt]{\textbf{Chain-of-Query (Ours)}} &
\shortstack[c]{\textbf{74.8} \\[-1pt] \footnotesize\textcolor{green!40!black}{\textbf{(+4.4)}}} &
\shortstack[c]{\textbf{92.3} \\[-1pt] \footnotesize\textcolor{green!40!black}{\textbf{(+4.7)}}} \\
\bottomrule
\end{tabular}
\label{tab:structured_methods}
\end{table}

\begin{table}[H]
\caption{Performance comparison of CoQ with baselines on the 57 challenging samples from WikiTQ identified by TabSQLify (GPT-3.5). CoQ achieves the highest accuracy, enabled by its schema abstraction, clause-by-clause SQL generation, and hybrid reasoning.}
\centering
\begin{tabular}{lc}
\toprule
\textbf{Method} & \textbf{Acc.} \\
\midrule
MAG-SQL & 15.79 \\
Chain-of-Table & 29.82 \\
StructGPT & 29.82 \\
Readi & 42.11 \\
H-STAR & \underline{47.37} \\
\hline
\\[-10pt]
\raisebox{0.9ex}[0pt]{\textbf{Chain-of-Query (Ours)}} &
\shortstack[c]{\textbf{57.89} \\[-1pt] \footnotesize\textcolor{green!40!black}{\textbf{(+10.52)}}} \\
\bottomrule
\end{tabular}
\label{tab:challenge_set}
\end{table}

\section{More Experimental Results}
\label{sec:appendix_more_experiments}

\subsection{Scalability to Stronger LLMs}
\label{sec:appendix_more_experiments_strong_llms}

Our initial experiments used GPT-3.5 and LLaMA 2 to ensure fair comparisons with baselines. To assess how CoQ scales with stronger models, we further evaluate it on LLaMA~3.1, DeepSeek-V3, and GPT-4.1 using the WikiTQ dataset. As shown in Table~\ref{tab:wikiq_sota_models}, CoQ consistently outperforms other methods regardless of model strength. Notably, even with GPT-3.5, CoQ achieves 74.77\% accuracy (Table~\ref{tab:performance_wiki_fact}), surpassing the performance of Few-Shot QA with GPT-4.1 (70.97\%). This demonstrates that our framework contributes more to overall performance than simply upgrading to a more powerful language model. When paired with GPT-4.1, CoQ further boosts accuracy to 84.92\% and reduces the invalid SQL rate to just 1.54\%, confirming that its combination of hybrid reasoning and robust SQL generation remains effective with stronger LLMs.

\begin{table}[H]
\caption{Error type distribution of CoQ on the challenging WikiTQ subset (GPT-3.5). Most remaining errors stem from incorrect reasoning or annotation noise.}
\centering
\renewcommand{\arraystretch}{0.95}
\setlength{\tabcolsep}{4pt}
\begin{tabular}{lcc}
\toprule
\textbf{Error Type} & \textbf{Count} & \textbf{Rate} \\
\midrule
Missing Columns & 0 & 0\% \\
Missing Rows & 6 & 25\% \\
Incorrect Reasoning & 11 & 46\% \\
Incorrect Annotation & 7 & 29\% \\
\Xhline{1.2pt}
\end{tabular}
\label{tab:subset_error_analysis}
\end{table}

\subsection{Comparison Against General Structured Data Reasoning Methods}
\label{sec:appendix_more_experiments_structure}

We further compare CoQ with four strong general structured-data reasoning baselines on the WikiTQ and TabFact datasets. The results are presented in Table~\ref{tab:structured_methods}. CoQ achieves substantial accuracy gains on both datasets. These baselines usually either rely on LLMs to filter tables as text, or restrict reasoning to shallow paths without supporting deeper operations. These approaches ultimately require the LLM to reason directly over tabular structures, where LLMs perform poorly. In contrast, CoQ explicitly controls query construction with clause-by-clause SQL validation and delegates mechanical operations (filtering, aggregation) to SQL, reducing reasoning burden on the LLM.

In summary, the experiments demonstrate that while there are strong representatives of the general structured data QA family, their limitations in table-specific reasoning lead to a consistent performance gap relative to CoQ. Interestingly, Readi and Chain-of-Knowledge cannot even directly support the TabFact benchmark. CoQ’s table-tailored design, which combines fine-grained SQL construction with hybrid reasoning, yields clear and stable performance advantages, particularly on more challenging questions.

\subsection{Performance on Challenging Subset and Error Analysis}
\label{sec:appendix_more_experiments_error}

\subsubsection{Results on Challenging WikiTQ Subset}

We evaluated our CoQ on the 57 error samples from WikiTQ provided by the GitHub repository\footnote{\url{https://github.com/mahadi-nahid/TabSQLify/blob/main/analysis/wtq_err_analysis_tabsqlify_binder_common.csv}} of TabSQLify \citep{nahid2024tabsqlifyenhancingreasoningcapabilities} and conducted a detailed error analysis. Table~\ref{tab:challenge_set} displays the results. CoQ correctly answers 33 out of 57 questions (57.89\%). In contrast, according to TabSQLify, both TabSQLify and BINDER fail to correctly answer any of these samples. In addition, CoQ achieves significantly larger margins on the challenging subset over other baselines. This confirms that our framework is not only competitive on the overall benchmark but also more robust on hard cases where existing methods struggle.

\subsubsection{Error Analysis of CoQ}

We further analyze the 24 samples in which CoQ produces incorrect answers. The error type information is shown in Table~\ref{tab:subset_error_analysis}. We adopt the error type definitions used in TabSQLify \citep{nahid2024tabsqlifyenhancingreasoningcapabilities} and H-STAR \citep{abhyankar2025hstarllmdrivenhybridsqltext}. In detail, "Missing Columns" and "Missing Rows" refer to missing necessary columns and rows, respectively. "Incorrect Reasoning" occurs when CoQ extracts the correct information, but LLM fails to produce the correct answer. "Incorrect Annotations" include semantically identical answers in different formats, ambiguous questions, and incorrect gold answers. Although the chosen subset is more challenging, CoQ still reduces errors in both sub-table retrieval and reasoning. And errors caused by "Incorrect Annotation" remain a notable fraction, which is analyzed below.

\subsubsection{Examples of "Incorrect Annotation"}

Following the definitions in TabSQLify and H-STAR, "Incorrect Annotation" covers semantically identical answers in different formats, ambiguous questions, and incorrect gold answers. We chose examples from the error cases above to explain the error type "Incorrect Annotation" in detail.

\noindent\textbf{1. Different Formats}

\begin{itemize}[noitemsep, topsep=1pt]
    \item Sample ID: nu-91
    \item Question: which country has the larger number of circuits?
    \item Answer by CoQ: united states
    \item Gold Answer: usa
    \item Explanation: Semantically identical answers in different formats.
\end{itemize}

\noindent\textbf{2. Ambiguous Questions}

\begin{figure}[htbp]
  \centering
  \includegraphics[width=\linewidth]{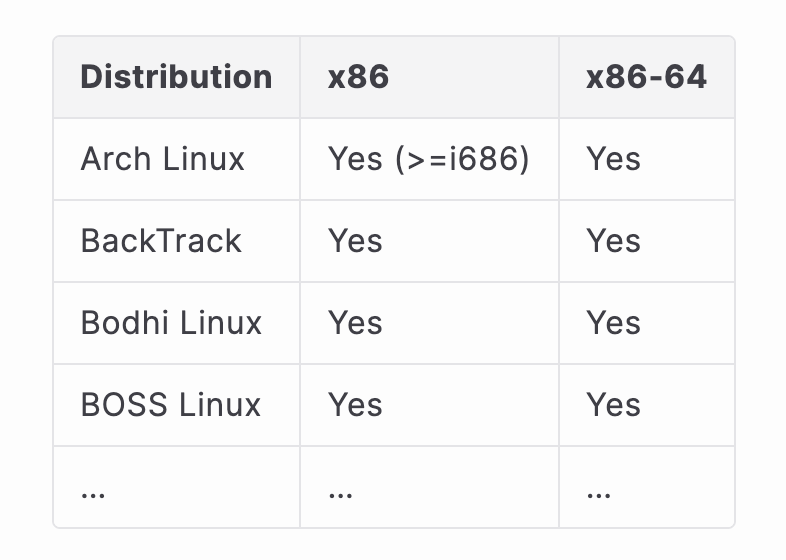}
  \caption{Partial Table of Sample nu-114.}
  \label{fig:case_ambiguous}
\end{figure}

\begin{itemize}[noitemsep, topsep=1pt]
    \item Sample ID: nu-114
    \item Question: how many distributions support the x86 architecture?
    \item Answer by CoQ: united states
    \item Gold Answer: usa
    \item Explanation: There are 29 rows in total having "Yes" in the column "x86", while the gold answer doesn't consider "Yes (>=i686)" shown in Figure~\ref{fig:case_ambiguous} as the distribution supporting x86. There is no constraint or clarification in the question.
\end{itemize}

\noindent\textbf{3. Incorrect Gold Answers}

\begin{itemize}[noitemsep, topsep=1pt]
    \item Sample ID: nu-0
    \item Question: which country had the most cyclists finish within the top 10?
    \item Answer by CoQ: spain, italy
    \item Gold Answer: italy
    \item Explanation: Both Spain and Italy have 3 cyclists in the top 10, but the gold answer only lists Italy.
\end{itemize}

\subsection{Ablation Analysis by Question Type}
\label{sec:appendix_more_experiments_ablation}

We conducted fine-grained ablations on the challenging 57-question subset of WikiTQ reported in Appendix~\ref{sec:appendix_more_experiments_error}. We labeled each question as Lookup, Aggregation, Comparison, Bridge, or Arithmetic, and compared the performance of CoQ and its ablated variants. The results are shown in Table \ref{tab:ablation_by_question}.

According to the result, Clause-by-clause SQL is particularly critical for aggregation and arithmetic questions (from 40.0\% to 6.7\% and from 80.0\% to 50.0\%), confirming its role in preventing error propagation across multi-step symbolic reasoning. Hybrid reasoning significantly benefits comparison and bridge questions (from 56.3\% to 25.0\% and from 57.1\% to 14.3\%), as SQL handles mechanical filtering/aggregation while LLMs focus on logical inference. Parallel decomposition provides the largest gains for comparison questions (from 56.3\% to 43.8\%), where independent sub-questions can be solved more efficiently in parallel. Natural-language schema improves robustness on lookup and aggregation questions by mitigating schema noise, though on some bridge questions noisy sampled values may occasionally mislead the model.

In short, this fine-grained analysis confirms that each module contributes in complementary ways: clause-by-clause SQL mitigates error propagation in multi-step symbolic reasoning, hybrid reasoning improves logical generalization, parallel decomposition accelerates comparison-type queries, and the natural-language schema stabilizes performance under irregular table structures. These results further demonstrate the necessity of integrating all four modules in CoQ to achieve robust table understanding.

\subsection{Step-wise LLM Call Breakdown and Empirical Token-level Cost}
\label{sec:appendix_more_experiments_cost}

First, we present a comparison of the theoretical number of LLM calls per question on WikiTQ, with a detailed step-wise breakdown shown in Table~\ref{tab:detailed_calls}. 

Given the variety and complexity of baseline frameworks, we argue that a token-level analysis, in addition to the theoretical number of LLM calls, provides a more comprehensive assessment of efficiency. Token-level processing volume offers a more accurate measure of computational cost and better reflects the overall efficiency of CoQ. 

Therefore, we conducted both token- and API-level cost analyses on the challenging WikiTQ subset described in Appendix~\ref{sec:appendix_more_experiments_error}. As this subset focuses on the most difficult questions, the measured cost represents a higher-than-average workload and thus provides an upper-bound estimate of token and resource consumption. The detailed results are presented in Table~\ref{tab:token_cost_analysis}. CoQ achieves significantly higher accuracy with only moderate additional expense, confirming that its fine-grained SQL generation and natural-language-style schema representation provide a scalable and effective balance between performance and efficiency.

\begin{table*}
\setlength{\tabcolsep}{2.3pt} 
\caption{Ablation results on the subset of 57 questions from WikiTQ (GPT-3.5), with breakdown by question type. CoQ achieves the highest accuracy through its modular design, with each component contributing significantly to overall performance.}
\begin{tabular}{lcccccc}
\toprule
\textbf{Method} & \textbf{Lookup} & \textbf{Aggregation} & \textbf{Comparison} & \textbf{Bridge} & \textbf{Arithmetic} & \textbf{Overall} \\
\midrule
\textsc{Chain-of-Query} & \textbf{100.0} & \textbf{40.0} & \textbf{56.3} & 57.1 & \textbf{80.0} & \textbf{57.9} \\
w/o Natural-Language-Schema & 50.0 & 26.7 & 50.0 & \textbf{64.3} & 60.0 & 49.1 \\
w/o Parallel Decomposition & 50.0 & 26.7 & 43.8 & 57.1 & 70.0 & 47.4 \\
w/o Clause-by-Clause Generation & 100.0 & 6.7 & 12.5 & 42.9 & 50.0 & 28.1 \\
w/o Hybrid Reasoning Division & 100.0 & 20.0 & 25.0 & 14.3 & 50.0 & 28.1 \\
\bottomrule
\end{tabular}
\label{tab:ablation_by_question}
\end{table*}

\renewcommand{\arraystretch}{0.9}
\setlength{\tabcolsep}{4pt}
\begin{table*}[ht]
\caption{Comparison of the theoretical LLM call counts per question (step-wise breakdown) across methods on the WikiTQ dataset (GPT-3.5).}
\centering
\begin{tabular}{lp{10cm}c}
\toprule
\textbf{Method} & \textbf{\# samples / step} & \textbf{Total \# samples} \\
\midrule
BINDER & Neural SQL: 50 & 50 \\\addlinespace[4pt]
DATER & Decompose Table: 40, Generate Cloze: 20, Generate SQL: 20, Answer: 20 & 100 \\\addlinespace[4pt]
Chain-of-Table & Dynamic Plan $\le$ 5, Generate Args $\le$ 19, Answer: 1 & $\le$ 25 \\\addlinespace[4pt]
Tree-of-Table & Dynamic Plan $\le$ 7, Generate Args $\le$ 21, Answer: 1 & $\le$ 29 \\\addlinespace[4pt]
MAC-SQL & Manager $\le$ 3, Selector $\le$ 3, SQL Generation $\le$ 3, SQL Correction $\le$ 9, Answer: 1 & $\le$ 19 \\\addlinespace[4pt]
MAG-SQL & Manager $\le$ 3, Summarization $\le$ 3, Selector $\le$ 3, SQL Generation $\le$ 6, SQL Correction $\le$ 9, Answer: 1 & $\le$ 25 \\\addlinespace[4pt]
TabSQLify & Table Decompose: 1--3, Answer: 1 & 2--4 \\\addlinespace[4pt]
H-STAR & Column Extraction: 2--4, Row Extraction: 2--4, Answer: 2 & 6--10 \\\addlinespace[4pt]
CoQ (Ours) & Semantic Splitter: 1--2, Dynamic Plan: 1--4, SELECT Clause: 1, WITH AS Clause with Correction: 0--3, WHERE Clause with Correction: 0--3, Aggregation Clause with Correction: 0--3, ORDER BY Clause with Correction: 0--3, Answer: 1--3 & 4--22 \\
\bottomrule
\end{tabular}
\label{tab:detailed_calls}
\end{table*}

\begin{table*}[ht]
\caption{Performance comparison of CoQ and strong baselines on the challenging WikiTQ subset (GPT-3.5), with detailed token-level cost statistics. "Acc." denotes accuracy (\%), "Avg Input/Output Token" is measured in thousands of tokens (k), and "Cost" represents the average USD (\$) per query. CoQ achieves the highest accuracy while maintaining competitive cost.}
\begin{tabular}{>{\centering\arraybackslash}m{4cm}
                >{\centering\arraybackslash}m{2cm}
                >{\centering\arraybackslash}m{2cm}
                >{\centering\arraybackslash}m{2cm}
                >{\centering\arraybackslash}m{2cm}
                >{\centering\arraybackslash}m{2cm}}
\toprule
\makecell{\textbf{Method}} &
\makecell{\textbf{Acc.}} &
\makecell{\textbf{Avg Input} \\[-2pt] \textbf{Token}} &
\makecell{\textbf{Avg Output} \\[-2pt] \textbf{Token}} &
\makecell{\textbf{Avg Input} \\[-2pt] \textbf{Cost}} &
\makecell{\textbf{Avg Output} \\[-2pt] \textbf{Cost}} \\
\midrule
MAG-SQL & 15.79 & \textbf{6.72} & \textbf{0.35} & \textbf{0.00336} & \textbf{0.00053} \\
QueryAgent & 17.54 & 13.84 & 0.58 & 0.00692 & 0.00087 \\
Chain-of-Table & 29.82 & 17.94 & 1.21 & 0.00897 & 0.00182 \\
StructGPT & 29.82 & 10.75 & 0.74 & 0.00538 & 0.00111 \\
Readi & 42.11 & \underline{7.82} & \underline{0.50} & \underline{0.00391} & \underline{0.00075} \\
H-STAR & \underline{47.37} & 9.10 & 1.18 & 0.00455 & 0.00177 \\
\textbf{Chain-of-Query (Ours)} & \textbf{57.89} & 10.62 & 0.91 & 0.00531 & 0.00137 \\
\bottomrule
\end{tabular}
\label{tab:token_cost_analysis}
\end{table*}

\clearpage

\raggedbottom

\section{Prompts Examples of Chain-of-Query}
\label{sec:appendix_prompts_coq}

Our prompts are uniform across datasets and not domain-specific.

\subsection{Prompt of Parallel Decomposition (WikiTQ)}

\begin{tcolorbox}[colback=gray!10, colframe=black, sharp corners=south, boxrule=0.8pt, width=\columnwidth, before skip=10pt, after skip=10pt]

\textbf{[Instruction]}\\
Your task is to decompose a question into subquestions.
The decomposition should be based on the presence of interrogative words (e.g., what, which, who, where, how, when, why) in the original question.
You should write the subquestions in the format of a python list.
Solve the task step by step if needed.

\medskip
\textbf{[Constraints]}\\
Subquestions should be independent and self-contained.
Avoid using pronouns like "he," "she," "it," or "they" if they refer to an entity introduced in another subquestion.
Restate the entity or essential context in each subquestion so that it makes complete sense on its own.
Ensure each subquestion is a complete, grammatically correct sentence.

\medskip
\textbf{[Response format]}\\
Your response should be in this format:
Analysis:
**[Your analysis]**
Subquestions:
```python
subquestions = [
    "first subquestion",
    "second subquestion",
    ...
]
```

\end{tcolorbox}

\subsection{Prompt of Sub-answer Generation (WikiTQ)}

\begin{tcolorbox}[colback=gray!10, colframe=black, sharp corners=south, boxrule=0.8pt, width=\columnwidth, before skip=10pt, after skip=10pt]

\textbf{[Instruction]}\\
Your task is to answer a question related to a given table based on the execution result attained by running SQLite.
Solve the task step by step if you need to.
The table schema, a few example rows, a SQLite query and the execution result will be provided.
Assume that you can always find the answer, so you must give an answer that makes sense to the question based on the given table.
Your answer should be as short as possible. Do not use sentences if one or two words will do.

\medskip
\textbf{[Response format]}\\
Your response should be in this format:
Analysis:
**[Your analysis]**
Answer:
[Your answer]

\end{tcolorbox}

\subsection{Prompt of Final Answer Generation (WikiTQ)}

\begin{tcolorbox}[colback=gray!10, colframe=black, sharp corners=south, boxrule=0.8pt, width=\columnwidth, before skip=10pt, after skip=10pt]

\textbf{[Instruction]}\\
Your task is to generate a final, coherent answer by combining the provided subanswers.
The final answer should fully and naturally address the original question, using information from all the subanswers.
Integrate the information fluently into a single, cohesive sentence, without explicitly listing or numbering the subanswers.
Write the final answer in clear, natural, and grammatically correct English.
Solve the task step by step if needed.
The original question, subquestions and subanswers will be provided.

\medskip
\textbf{[Constraints]}\\
Use all the subanswers when forming the final answer.
Do not simply list the subanswers separately; integrate them fluently into a single, cohesive sentence.
If an entity is introduced clearly in one subanswer, you may refer back to it later using appropriate pronouns such as "he," "she," "it," or "they."
Avoid repeating the full name or full description unnecessarily when a pronoun would maintain clarity and improve fluency.
Ensure that the final answer sounds fluent and directly addresses the original question.
If any subanswers overlap or feel redundant, merge and rephrase them appropriately.
If relevant, you may keep specific numbers or data in your final answer to make it more informative, even if the numbers are not explicitly asked for in the question.
Avoid run-on or grammatically incorrect structures even though the answer should be a single sentence.

\medskip
\textbf{[Response format]}\\
Your response should be in this format:
Analysis:
**[Your analysis]**
Answer:
[Your answer]

\end{tcolorbox}

\subsection{Prompt of SELECT-FROM Clause Generation}

\begin{tcolorbox}[colback=gray!10, colframe=black, sharp corners=south, boxrule=0.8pt, width=\columnwidth, before skip=10pt, after skip=10pt]

\textbf{[Instruction]}\\
Your task is to fill in the missing column names in an incomplete SQLite query so that it extracts the columns required to interpret the question correctly.
Solve the task step by step if you need to.
The table schema, a few example rows, question, and an incomplete SQLite query will be provided.

\medskip
\textbf{[Constraints]}\\
The SQLite does not need to directly answer the question.
You must complete the SELECT clause so it contains:
1. All value columns relevant to the question. 2. All context columns needed to understand what each row represents (e.g., identifiers, categories, descriptions).
Only insert the missing column names in the parentheses of the SELECT statement without modifying any other part of the given SQL query.
Don't add other clauses like WHERE and GROUP BY.
Always include context columns like description, name, category, or type when they are essential to interpreting the selected values.

\medskip
\textbf{[Response format]}\\
Your response should be in this format:
Analysis:
**[Your analysis]**
SQL:
```sql
[the completed SQL]
```

\end{tcolorbox}

\subsection{Prompt of WHERE Clause Generation}

\begin{tcolorbox}[colback=gray!10, colframe=black, sharp corners=south, boxrule=0.8pt, width=\columnwidth, before skip=10pt, after skip=10pt]

\textbf{[Instruction]}\\
Your task is to fill in the WHERE clause in an incomplete SQLite query so that it extracts a useful subset of rows that are most relevant to answering the question, even if the query is not final.
Solve the task step by step if you need to.
The table schema, a few example rows, question, and an incomplete SQLite query will be provided.

\medskip
\textbf{[Constraints]}\\
The SQLite does not need to directly answer the question.
Insert the necessary condition(s) or subquery in the WHERE clause, do not modify any other part of the provided SQL query.
Use relaxed fuzzy matching (e.g., LIKE, IN) as much as possible when unsure of exact values.
When filtering based on a value from the question (e.g., a team, location, or result), ensure the corresponding column contains similar values that match the target entity.
Avoid filtering on columns when it is unclear if they support the value mentioned in the question. Do not assume a column's semantics solely based on its name.
Do not invent or hallucinate values unless they are clearly implied by the question and supported by the schema.
Do not add additional SQL clauses such as GROUP BY or ORDER BY.

\medskip
\textbf{[Response format]}\\
Your response should be in this format:
Analysis:
**[Your analysis]**
SQL:
```sql
[the completed SQL]
```

\end{tcolorbox}

\clearpage

\subsection{Prompt of WITH AS Clause Generation}

\begin{tcolorbox}[colback=gray!10, colframe=black, sharp corners=south, boxrule=0.8pt, width=\columnwidth, before skip=10pt, after skip=10pt]

\textbf{[Instruction]}\\
Your task is to write a WITH ... AS SELECT statement that restructures the original table to enable accurate computation or aggregation (e.g., summing, grouping). These computations are needed to answer a provided question. You should extract or combine relevant parts of one or more columns selected in the provided SQLite query to create new columns in the new table.
Solve the task step by step if you need to.
The original table schema, a few example rows, the question, and a basic SQLite query (choosing one or more columns) will be provided.

\medskip
\textbf{[Constraints]}\\
The question requires an operation (such as summing numeric values) that cannot be performed correctly without splitting or concatenating current column(s).
Use SQLite syntax and functions (such as SUBSTR, INSTR, or string concatenation) as needed.
Be mindful of data types in the SELECT subquery to ensure correct comparisons, computations, and transformations.
Preserve all columns selected by the provided SQLite query in the new table, even if they do not require transformation.

\medskip
\textbf{[Response format]}\\
Your response should be in this format:
Analysis:
**[Your analysis]**
SQL:
```sql
[the completed SQL]
```

\end{tcolorbox}

\subsection{Prompt of Aggregate Function Clause Generation}

\begin{tcolorbox}[colback=gray!10, colframe=black, sharp corners=south, boxrule=0.8pt, width=\columnwidth, before skip=10pt, after skip=10pt]

\textbf{[Instruction]}\\
Your task is to choose exactly one aggregate function from the list (COUNT, AVG, MAX, MIN, SUM) and rewrite the provided SQLite query. This query retrieves information from a table to answer a given question.
Solve the task step by step if you need to.
The table schema, a few example rows, question, and a basic SQLite query will be provided.

\medskip
\textbf{[Constraints]}\\
The SQLite does not need to directly answer the question.
Only consider following aggregate functions: COUNT, AVG, MAX, MIN, SUM.
Aggregate functions must not be nested.
The aggregate function should be chosen based on the semantics of the question (e.g., "total" suggests SUM, "most" suggests MAX).
Modify the basic query by applying the chosen function to the relevant column.
Keep the structure of the original query intact, only apply aggregation where needed.
Use aggregate functions properly over the whole table.

\medskip
\textbf{[Response format]}\\
Your response should be in this format:
Analysis:
**[Your analysis]**
SQL:
```sql
[the completed SQL]
```

\end{tcolorbox}

\clearpage

\subsection{Prompt of ORDER BY Clause Generation}

\begin{tcolorbox}[colback=gray!10, colframe=black, sharp corners=south, boxrule=0.8pt, width=\columnwidth, before skip=10pt, after skip=10pt]

\textbf{[Instruction]}\\
Your task is to add an ORDER BY clause to the provided SQLite query. This query retrieves information from a table to answer a given question.
Solve the task step by step if you need to.
The table schema, a few example rows, question, and a basic SQLite query will be provided.

\medskip
\textbf{[Constraints]}\\
The SQLite does not need to directly answer the question.
Modify the query by applying an ORDER BY clause to the relevant column.
The sorting order (ASC or DESC) should be determined based on the semantics of the question.
Apply the ORDER BY clause to the relevant column.
Ensure that the original query structure remains unchanged except for the addition of the ORDER BY clause.

\medskip
\textbf{[Response format]}\\
Your response should be in this format:
Analysis:
**[Your analysis]**
SQL:
```sql
[the completed SQL]
```

\end{tcolorbox}

\subsection{Prompt of Dynamic Planner (Sufficiency)}

\begin{tcolorbox}[colback=gray!10, colframe=black, sharp corners=south, boxrule=0.8pt, width=\columnwidth, before skip=10pt, after skip=10pt]

\textbf{[Instruction]}\\
Your task is to decide whether the current SQL result is sufficient to answer the question.
The SQLite query retrieves information from a table to answer a given question.
Solve the task step by step if needed.
The table schema, a few example rows, the question, a SQLite query and its result will be provided.

\medskip
\textbf{[Constraints]}\\
Answer "Yes" if and only if the current SQL result is sufficient to answer the question or with reasonable interpretation.
Otherwise, answer "No".
Base your decision solely on the given question and the provided basic SQLite query.
The SQLite query does not need to be a complete or final answer. If it includes most key information that allows someone to infer the answer correctly, "Yes" is still acceptable, even if the question suggests that additional filtering, aggregation, or sorting might be needed.

\medskip
\textbf{[Response format]}\\
Your response should be in this format:
Analysis:
**[Your analysis]**
Decision:
[Yes or No]

\end{tcolorbox}

\subsection{Prompt of Dynamic Planner (WHERE Clause)}

\begin{tcolorbox}[colback=gray!10, colframe=black, sharp corners=south, boxrule=0.8pt, width=\columnwidth, before skip=10pt, after skip=10pt]

\textbf{[Instruction]}\\
Your task is to decide whether answering the question require using a SQL WHERE clause to filter rows in the table?
Solve the task step by step if you need to.
The table schema, the total number of rows, a few example rows, and the question will be provided.

\medskip
\textbf{[Constraints]}\\
Filtering must use a WHERE clause. Don't consider other filtering methods like GROUP BY.
Check the number of rows first. If the question refers to a subset like “top 10,” but the entire table already contains only that subset (e.g., the table has exactly 10 rows), then no filtering is needed and the answer is No.
Operations like ORDER BY, LIMIT, or aggregation across all rows do not count as filtering.
You must make the decision.
Output Yes or No as the decision.

\medskip
\textbf{[Response format]}\\
Your response should be in this format:
Analysis:
**[Your analysis]**
Decision:
[Yes or No]

\end{tcolorbox}

\subsection{Prompt of WHERE Clause Correction}
\label{appendix_where}

\begin{tcolorbox}[colback=gray!10, colframe=black, sharp corners=south, boxrule=0.8pt, width=\columnwidth, before skip=10pt, after skip=10pt]

\textbf{[Instruction]}\\
When executing the SQLite query below, an error occurred in the WHERE clause. Please correct the WHERE clause based on the provided question, table schema and error information.
Solve the task step by step if needed. Identify which conditions are irrelevant, overly strict, or mismatched.
The corrected WHERE clause should retrieve rows that can help interpret the question accurately, even if loosely filtered.
The table schema, a few example rows, question, the incorrect SQLite query and error information will be provided.
Only fix the WHERE clause, do not change any other part of the query.
After fixing the query, verify it carefully. If possible, include verifiable evidence in your analysis.

\medskip
\textbf{[Constraints]}\\
Modify only the WHERE clause; do not alter SELECT, GROUP BY, or other clauses.
Ensure that the condition(s) in WHERE match the logic required by the question.
Use correct syntax and valid condition(s) based on the provided schema.
Handle data types appropriately. e.g., avoid comparing numeric strings without conversion.
Use relaxed fuzzy matching (e.g., LIKE, IN) as much as possible when unsure of exact values.
If the SQL query might return no results because of overly strict or irrelevant conditions, modify it to be more flexible:
- Remove conditions when value mismatches are likely.
- Relax LIKE or inequality comparisons that don't match the table's value patterns.
- Do not remove constraints that clearly align with the question.
Remember: a runnable and interpretable query is better than an overly strict query that returns nothing.

\medskip
\textbf{[Response format]}\\
Your response should be in this format:
Analysis:
**[Your analysis]**
SQL:
```sql
[the correct SQL]
```

\end{tcolorbox}

\section{Prompt Example of Basic Text-to-SQL (WikiTQ)}
\label{sec:appendix_prompts_bt2sql}

\begin{tcolorbox}[colback=gray!10, colframe=black, sharp corners=south, boxrule=0.8pt, width=\columnwidth, before skip=10pt, after skip=10pt]

\textbf{[Instruction]}\\
Given the table schema and three example rows out of the table, write a SQLite program to extract the sub-table that contains the information needed to answer the question.
The SQLite does not need to directly answer the question.
Assume you always have enough information when executing the SQLite.
Output only the SQL, with no explanation.

\medskip
\textbf{[Response format]}\\
Your response should be in this format:
SQL:
```sql
[the completed SQL]
```

\end{tcolorbox}

\section{Prompt Example of End-to-End QA (WikiTQ)}
\label{sec:appendix_prompts_e2eqa}

\begin{tcolorbox}[colback=gray!10, colframe=black, sharp corners=south, boxrule=0.8pt, width=\columnwidth, before skip=10pt, after skip=10pt]

\textbf{[Instruction]}\\
Your task is to answer a question related to a given table.
Assume you can always find the answer, so you must give an answer that makes sense to the question based on the given table.
Your answer should be as short as possible. Do not use sentences if one or two words will do.
Output only the answer, with no explanation.

\end{tcolorbox}

\clearpage

\end{document}